\pdfoutput=1

\documentclass[11pt]{article}

\usepackage[preprint]{acl}

\usepackage{times}
\usepackage{latexsym}
\usepackage{array}
\usepackage{arydshln}
\usepackage{tikz}
\usepackage{floatpag}
\usepackage{makecell}
\usepackage{booktabs}
\usepackage{multirow}
\usepackage{subfig}
\usepackage{stackengine}
\usepackage[]{graphicx}

\usepackage[T1]{fontenc}

\usepackage[utf8]{inputenc}
\usepackage{csquotes}
\usepackage{microtype}

\usepackage{inconsolata}

\title{The Inherent Limits of Pretrained LLMs: The Unexpected Convergence of Instruction Tuning and In-Context Learning Capabilities}

\author{
  \textbf{Irina Bigoulaeva\textsuperscript{1}},
  \textbf{Harish Tayyar Madabushi\textsuperscript{2}} 
  \textbf{Iryna Gurevych\textsuperscript{1}} \\
  \textsuperscript{1}Ubiquitous Knowledge Processing Lab, Technical University of Darmstadt \\
  \textsuperscript{2}Department of Computer Science, The University of Bath \\
  \texttt{\small www.ukp.tu-darmstadt.de} \\ [-0.1mm]
\texttt{\small htm43@bath.ac.uk} \\ }

\begin{document}
\maketitle
\begin{abstract}
Large Language Models (LLMs), trained on extensive web-scale corpora, have demonstrated remarkable abilities across diverse tasks, especially as they are scaled up. Nevertheless, even state-of-the-art models struggle in certain cases, sometimes failing at problems solvable by young children, indicating that traditional notions of task complexity are insufficient for explaining LLM capabilities. However, exploring LLM capabilities is complicated by the fact that most widely-used models are also `instruction-tuned' to respond appropriately to prompts. With the goal of disentangling the factors influencing LLM performance, we investigate whether instruction-tuned models possess fundamentally different capabilities from base models that are prompted using in-context examples. Through extensive experiments across various model families, scales and task types, which included instruction tuning 90 different LLMs, we demonstrate that the performance of instruction-tuned models is significantly correlated with the in-context performance of their base counterparts. By clarifying what instruction-tuning contributes, we extend prior research into in-context learning, which suggests that base models use priors from pretraining data to solve tasks. Specifically, we extend this understanding to instruction-tuned models, suggesting that their pretraining data similarly sets a limiting boundary on the tasks they can solve, with the added influence of the instruction-tuning dataset.\footnote{\url{https://github.com/UKPLab/arxiv2025-inherent-limits-plms}}
\end{abstract}

\section{Introduction, Motivation and Context}
\label{sec:intro}
\begin{figure*}[htp]
    \centering
\stackunder[0pt]{\includegraphics[width=0.8\linewidth,trim={1cm 5cm 0cm 7cm},clip]{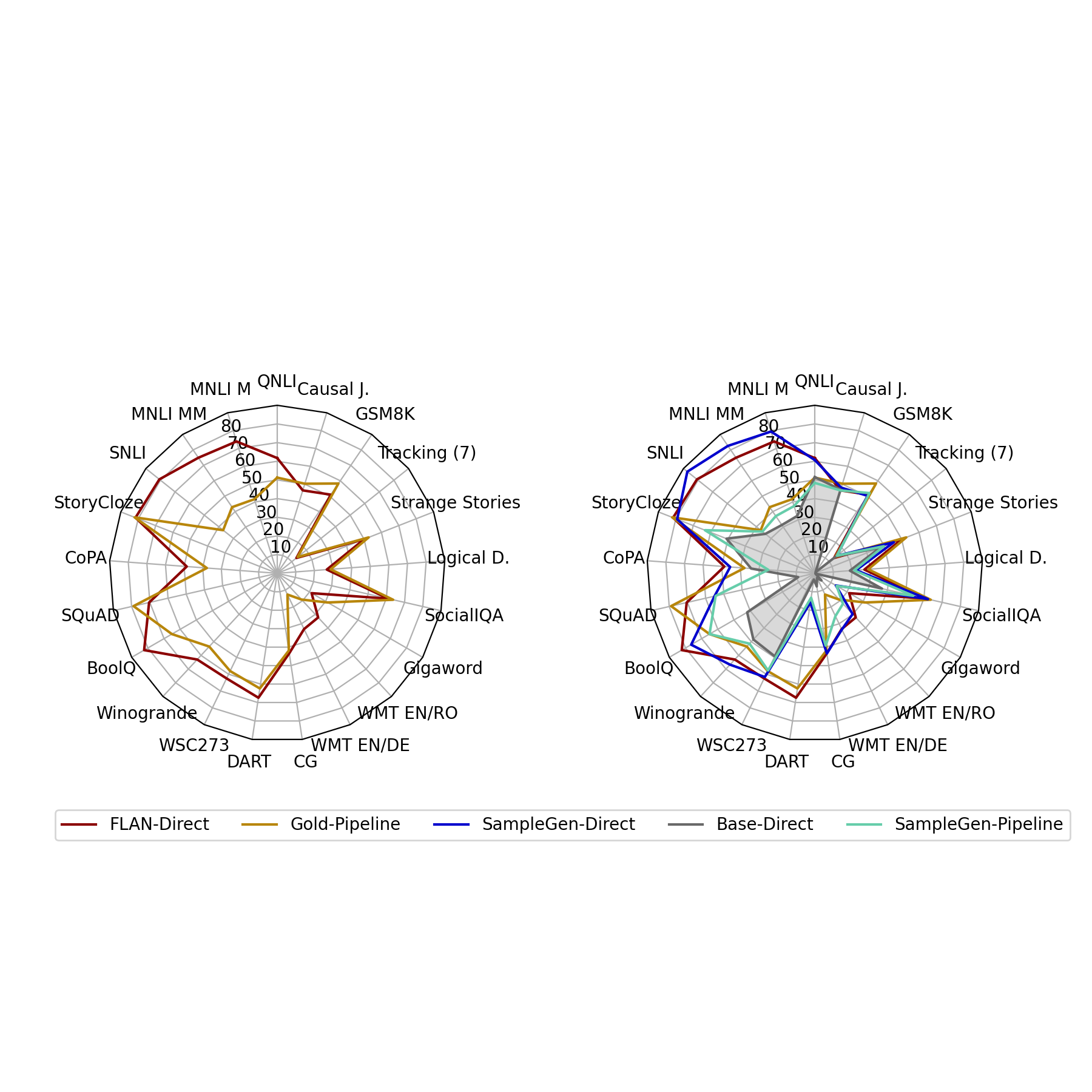}}{}%
\stackunder[0pt]{\includegraphics[width=0.5\linewidth,trim={0cm 7cm 15cm 0cm},clip]{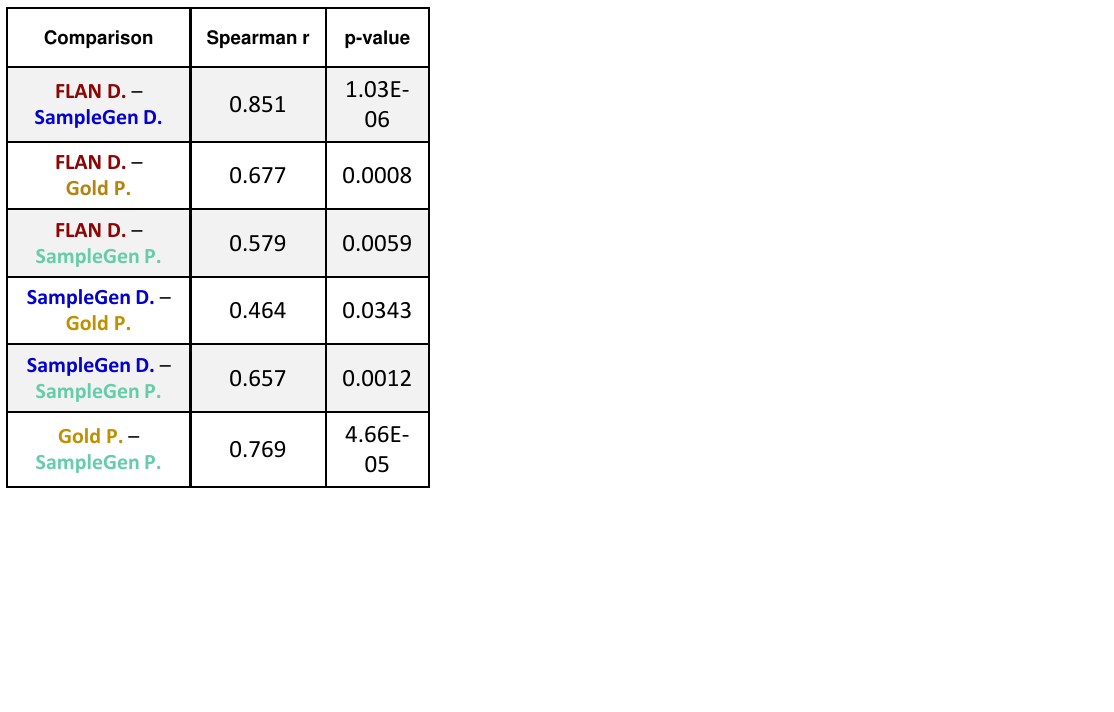}}{}

    \caption{Our main results for LLaMA-2 13B demonstrate a significant correlation between the performance of base models prompted with in-context examples (and their variants) and that of an instruction-tuned model. Base-D is the zero-shot base model, close to the random baseline. This correlation is additionally dependant on the task clusters present in the instruction-tuning data. This is particularly visible in the isolated comparison of the FLAN and Gold Pipeline models (left). Plots showing generalisation across model families and scale are in the Appendix (\ref{app:fig:llama_7B_hyp1}, \ref{app:fig:mistral_7B_hyp1}). 
    }
    \label{fig:main_charts}
\end{figure*}

Large Language Models (LLMs), trained primarily on the task of sequence completion on web-scale corpora, are now capable of solving a wide variety of tasks and interacting with us in meaningful ways through natural-language text. The set of tasks on which an LLM performs significantly above the random baseline in the zero-shot setting, i.e., \textit{task solvability}, is routinely used to compare models' capabilities~\cite{openai2024gpt4technicalreport,srivastava2023beyond,wei2022emergent,palm-paper}, and it has been shown that increasing the parameter count and the amount of training data of these models (i.e., \enquote{scaling up}) increases the range of tasks they can solve \cite{kaplan2020scalinglawsneurallanguage,palm-paper}. Some works suggest that LLMs can solve complex tasks, even in scenarios where they cannot rely on memory, for which humans would typically resort to reasoning~\cite{lu-etal-2024-emergent,srivastava2023beyond,wei2022emergent}.

However, despite their many capabilities, LLMs are not infallible. Even at the largest of scales, there remain tasks that LLMs cannot solve. For example, the most advanced models currently available, including GPT-4~\cite{openai2024gpt4technicalreport} and GPT-o1~\cite{openai2024gpt4ocard}, continue to fail on tasks that require reasoning and planning~\cite{valmeekam2024llmscantplanlrms,valmeekam2024planningstrawberryfieldsevaluating}. Most previous works have, either explicitly or implicitly, categorized the hardness of tasks for LLMs in light of their hardness for humans~\cite{srivastava2023beyond,wei2022emergent}. Interestingly, however, there is little correlation between the array of tasks that LLMs can solve and the array of tasks that can be solved by humans at various stages of cognitive development (e.g. childhood vs. adulthood). While LLMs are capable of solving advanced college-level science questions~\cite{palm-paper}, they often fail at tasks that seven- and ten-year-old children can solve with ease~\cite{shapira-etal-2023-well}. It is therefore clear that traditional, human-centric notions of `task complexity' fail to explain the success and failure of LLMs in solving tasks. In other words, \textit{we cannot reliably use information about how hard a task would be for a human to infer how difficult it would be for an LLM.} Thus, it becomes crucial to investigate which particular properties in fact influence LLM performance on that task. 

\subsection{Priors from Pretraining Data}

A likely source of this disconnect between human-centric task complexity and LLM performance is the possibility that LLMs use priors from pretraining data, just as priors from training data are central to the performance of any machine learning model.

However, systematically exploring such capabilities of LLMs is complicated by various post-training processes that are commonly employed to enhance LLMs after pretraining. Most notable is 
instructional fine-tuning (also known as instruction tuning), which improves understanding of user instructions~\cite{wei2022finetuned}. %
Instruction tuning is particularly important to disentangle, as models which are not instruction-tuned tend not to be able to directly solve complex tasks~\cite{lu-etal-2024-emergent}.

Prior work has demonstrated that base LLMs are likely to be using priors from pre-training data to solve tasks (detailed discussion in Section \ref{sec:base-icl}). In fact, instruction-tuned models struggle with ``counterfactual'' tasks—those that deviate from standard assumptions but rely on similar logic, such as performing addition in base-9%
~\cite{wu-etal-2024-reasoning}. We contend that this suggests that instruction-tuned models might similarly rely on priors from pretraining data, and focus on establishing if instruction tuning fundamentally alters what LLMs are capable of.

\subsection{Contributions}
Some studies claim that LLMs show early signs of `intelligence’ or `creativity’~\cite{bubeck2023sparks}, while others highlight their reliance on autoregressive mechanisms---the ability to predict the next word in a sequence~\cite{doi:10.1073/pnas.2322420121}. \emph{However, we argue that these abilities are better understood as a product of generalisation from priors from the massive datasets they were pretrained on. Crucially, these models require instruction tuning to handle complex tasks in the zero-shot setting without additional training, which raises an important question: What exactly does instruction tuning achieve?} Therefore, this work explores whether instruction tuning provides models with fundamentally new abilities or simply improves their ability to interpret instructions (see also Section \ref{sec:relatedinstructiontuning}). By answering this, we aim to clarify whether instruction-tuned models are free from the limitations of their pretraining when solving tasks or are limited to generalising from priors in their pretraining data, in the same way that their base counterparts are.
To this end, this work is focused on answering the following research question:
\begin{quote}
    To what extent do the performance and generalisability of an instruction-tuned LLM correlate with that of its corresponding base model across a diverse set of tasks, and do the models behave as expected when accounting for potentially confounding factors such as prompt complexity and tasks included in instruction tuning data? 
\end{quote} 

By carefully experimenting across models of different scales and families, we demonstrate that the performance of an instruction-tuned model on a task is significantly correlated with that of the corresponding base model, with the added influence of the instruction-tuning dataset. Furthermore, we find that our conclusions hold even when controlling for confounding factors. %
Since the pretraining datasets of the base and the instruction-tuned versions of the same model are identical, and that despite the additional instruction-tuning, the models' capabilities remain correlated, we can conclude that instruction-tuned LLMs are similarly limited as base models. This experimental setup allows us to evaluate LLMs whose pretraining data is not public.%

Demonstrating this correlation allows us to extend previous work, including work that pretrained models on carefully curated datasets, which indicates that the capabilities of base LLMs on specific tasks are linked to the existence of priors relevant to that task in pretraining data~\cite{wu-etal-2024-reasoning} (See also Section \ref{sec:base-icl}). In turn, we demonstrate that the performance of instruction-tuned models is similarly influenced by the existence of relevant priors in pretraining data. In other words, just as base models are limited by their pretraining data, so are instruction-tuned models. Regardless of the precise nature of this limitation, our work enables a more straightforward examination of what these models can achieve by focusing on the simpler base models, thanks to their single training objective. 
Importantly, following \citet{lu-etal-2024-emergent}, we do not claim that LLMs are only capable of reproducing information in the manner of memory recall; instead, they extrapolate from pretraining data.

\section{Context and Related Work}
In this section, we outline related work that provides the contextual foundation for our research.

\subsection{Base Models and In-Context Learning}
\label{sec:base-icl}
Prior research has explored whether LLMs are able to solve tasks that they were \textit{not} explicitly trained on. For example, research has observed so-called ``emergent abilities'' in LLMs, which seem to suggest that LLMs might indeed be capable of more than mere extrapolation from pretraining data \cite{wei2022emergent}. To the best of our knowledge, however, this is the first work to systematically explore the correlation between instruction-tuned models' ability to solve tasks and that of their base counterparts. 
More broadly, our work extends recent research on in-context learning in base models, demonstrating that instruction-tuned models have fundamentally similar limitations on the range of tasks they can effectively solve, possibly determined by their pretraining data. We detail this relevant prior work below. %

LLMs which have not undergone any post-training, referred to as \textit{base models} are not inherently capable of solving complex tasks that require more than memory retrieval~\cite{lu-etal-2024-emergent}.  However, if a base model is sufficiently large and is presented with examples of the task in the prompt, then the model will be able to generalise from these examples to respond to a new question related to that task~\cite{NEURIPS2020_1457c0d6}. This ability to generalise from a few examples to a unseen example is referred to as In-Context Learning (ICL). \footnote{We differentiate this from the ability of instruction-tuned models to leverage contextual information for task-solving, which some other works also describe as in-context learning.}%

The exact mechanism by which base LLMs generalise from examples in a prompt remains an open question. However, a growing body of research is beginning to explore this issue. Several works suggest that LLMs are capable of using certain mechanisms to perform meta-learning, which can be thought of as a form of gradient descent, to solve tasks in-context~\cite{JMLR:v25:23-1042,ahn2023transformers,dai-etal-2023-gpt}, while others have found contradictory evidence~\cite{deutch-etal-2024-context}. We present a more detailed description of ICL in Appendix \ref{app:icl}. While the exact mechanism for how LLMs perform ICL is still under exploration, what all these works have in common is that ICL is a form of extrapolation from pretraining data.

\subsection{Instruction Tuning}
\label{sec:relatedinstructiontuning}
In contrast to base models, the mechanisms driving the performance of instruction-tuned models remain poorly understood, largely due to the added complexity of the instruction-tuning process \cite{lu-etal-2024-emergent}. There is significant debate about the capabilities of these models: Some argue that instruction-tuned models exhibit fundamentally different behaviours from base models, including potentially deceptive capabilities~\cite{doi:10.1073/pnas.2317967121}, while others contend that their abilities do not go far beyond those of base models~\cite{lu-etal-2024-emergent}. 

While there exists prior work suggesting that instruction-tuning does not add new knowledge to models but builds on existing pretraining~\cite{zhou-lima-superficial, ghosh2024closerlooklimitationsinstruction}, our work is fundamentally different. Our study examines base and instruction-tuned models from the perspective of task solvability, comparing the types of tasks each kind of model can address, while prompting base models with in-context examples known to allow them to perform on complex tasks~\cite{lu-etal-2024-emergent}. 

Notably, prior research has shown that instruction-tuned LLMs struggle with `counterfactual' tasks~\cite{wu-etal-2024-reasoning}. This suggests that instruction-tuned LLMs may, like their base counterparts, also depend on patterns or priors available in their pretraining data, in a way that is not fundamentally altered by the instruction-tuning process.

\subsection{Emergent Abilities in LLMs}
Recent works have observed that certain LLMs develop `emergent abilities'---the ability to solve unseen tasks at a level far exceeding smaller models---despite lacking explicit training on those tasks \cite{wei2022emergent}.
Importantly, many of these emergent abilities are related to reasoning, which is a behaviour fundamentally different from extrapolation from pretraining data. However, various other works have called the existence of emergent abilities into question.~\citet{schaeffer2023are} and ~\citet{lu-etal-2024-emergent} suggest that there are simpler explanations, including in-context learning, to explain model performance. %
Our work differs from prior studies by systematically investigating the correlation between instruction-tuned models, which are said to exhibit emergent abilities, and base models, which are not~\cite{wei2022emergent,wei2022finetuned}. 
Our findings suggest that increased scale increases the ability of models to extract priors from pretraining data rather than leading to `emergent' behaviour.

\section{Experimental Setup}
In this section, we outline the general methods used to address the research question posed in Section \ref{sec:intro}. Details of the experiments and results are presented in Section \ref{sec:rq1}, followed by our experimental evaluation of biases in Section \ref{sec:rq2}.

\subsection{Comparing Base and Instruction-Tuned Models}
As discussed, since base LLMs cannot solve certain complex tasks~\cite{lu-etal-2024-emergent} without in-context examples, we evaluate their performance using the few-shot setting. When comparing the performance of base LLMs to that of instruction-tuned LLMs, it is important to account for the difference in their training data: Base LLMs see only a few in-context examples relevant to the target task, while instruction-tuned models have seen significantly larger amounts of data by virtue of being instruction-tuned. For example, instruction-tuning on one NLI task is likely to help with inference on other NLI tasks. For this reason, our primary aim is to identify the correlation between their performances, rather than make a direct (unfair) comparison. While these experiments allow us to assess the relationship between task solvability in base and instruction tuned models, it is also important to evaluate the \emph{generalisability} of base and instruction-tuned models, as discussed in the following section.

\subsubsection{Comparing Generalisability} 
A comparison between the numerical performance scores of base models and instruction-tuned models would be insufficient, as instruction-tuned models have seen much more additional data via instruction-tuning than is typically made available in-context to base models, and would therefore trivially have higher scores. In addition, instruction-tuned models are known to generalise to tasks not present in their instruction-tuning data. Thus, in addition to establishing a direct correlation between their performance, we must also evaluate the extent to which instruction-tuning itself allows models to \textit{generalise instructional understanding}. Specifically, if we instruction-tune a model to understand instructions pertaining to one task, will this generalise to the model understanding instructions pertaining to novel tasks? Moreover, how does this compare to a base model's ability to generalise when provided with in-context examples?

To answer these questions, we use a novel approach that involves instruction-tuning LLMs to generate task-specific examples (in-context examples) alongside the final answer. We refer to a model fine-tuned in this manner as \textsc{SampleGen}. \emph{These experiments aim to separate generalisability from task-solving abilities and evaluate them independently.} Crucially, to robustly establish the relationship between the capabilities of a base LLM and its instruction-tuned counterpart, we must also explore the correlation between their performance and generalisation abilities.

It is important to note that \textsc{SampleGen}'s final answer to the prompt might be entirely independent of the examples it generates prior to the answer, as we have no way of ensuring faithfulness. Consequently, evaluating its performance does not necessarily reflect the quality or generalisability of the examples it produces. To address this, we also evaluate a base LLM using the in-context examples generated by \textsc{SampleGen}. This evaluation pipeline is referred to as the \textsc{SampleGen Pipeline}. An overview of all the variations we use is persented in Table \ref{tab:evaltypes}.

\begin{table}[htp]
\footnotesize
\renewcommand{\arraystretch}{1.5}
\begin{center}
\begin{tabular}{ | m{1.3cm} | m{5.3cm}| }
\hline
Name & Description  \\
\hline
Base Direct & The base version of the model, without in-context examples. Since models typically cannot solve tasks in this configuration, it serves as a control. \\
Gold-Pipeline & The base model is prompted with gold in-context examples drawn from the corresponding training split of the evaluated task. \\
FLAN Direct & The zero-shot setting of the modified FLAN instruction-tuned version of the model, created by applying instruction tuning using the FLAN templates. \\
SampleGen Direct & The zero-shot setting of the SampleGen instruction-tuned version of the model, trained to generate task-specific examples alongside the answer to the given prompt. \\
SampleGen-Pipeline & The base model is prompted with in-context examples generated by the SampleGen model when provided with the evaluated task. \\
\hline
\end{tabular}
\end{center}
\caption{\label{tab:evaltypes} Descriptions of various training and evaluation settings used to compare base LLMs and instruction-tuned models, highlighting the use of in-context examples, instruction tuning, and task-specific pipelines.}
\end{table}

\subsection{Instruction Tuning and Evaluation Datasets}
\label{sec:instruction-tuning-and-evaluation-datasets}
\begin{figure*}
    \centering
    \includegraphics[width=\linewidth,scale=0.9,trim={0 2.5cm 0 0.5cm},clip]{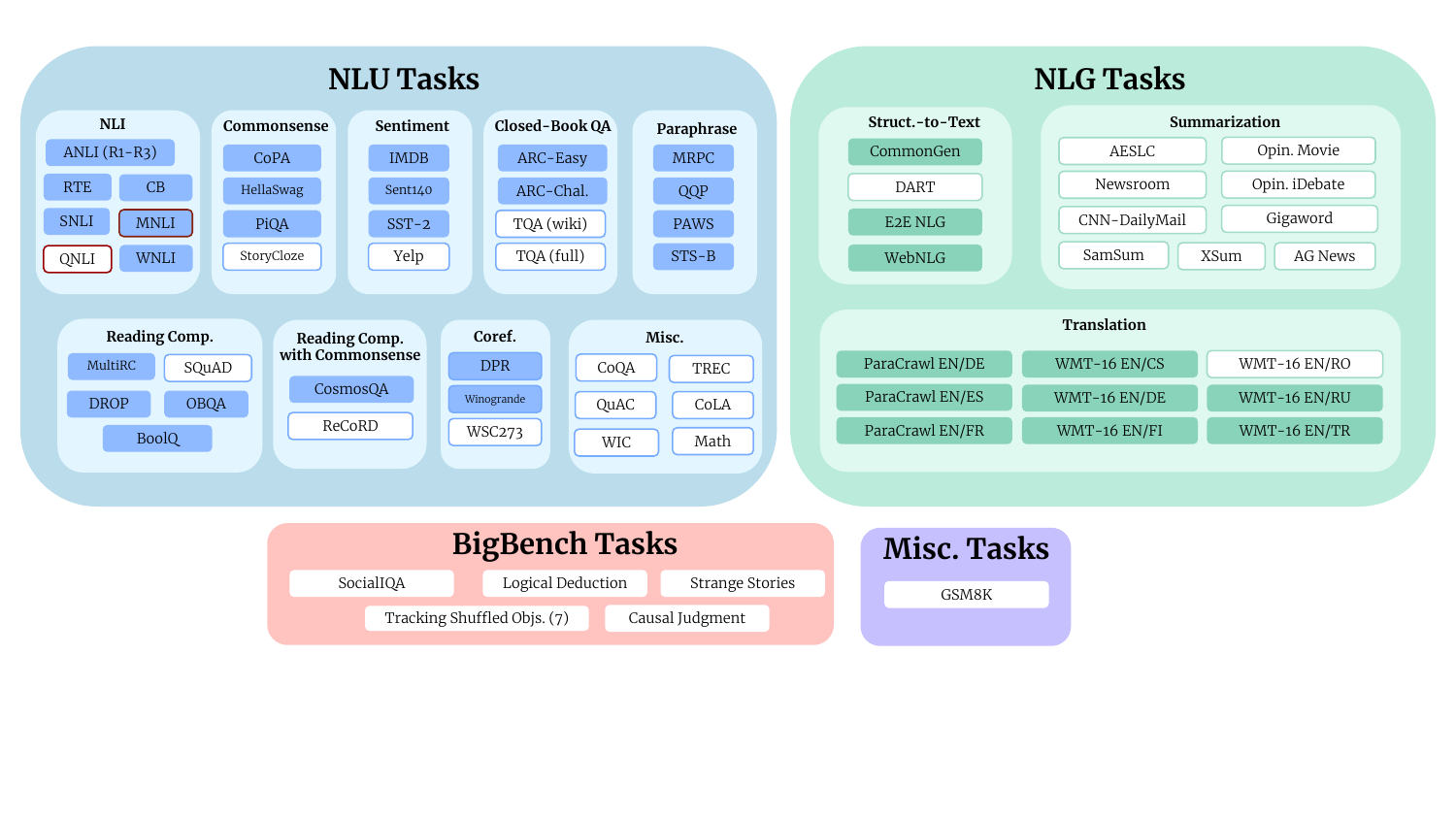}
    \caption{The tasks used in our experiments, grouped into clusters following \cite{wei2022finetuned}. Dark-colour backgrounds indicate that the training set of the corresponding task was used in instruction-tuning our models. The tasks with a white background were not seen during training. Tasks with a red border are those used for testing in the supplementary experiments in Section \ref{subsec:instruction-tuning-tasks}.}
    \label{fig:flan_datasets}
\end{figure*}

To ensure an effective and controlled comparison of base and instruction-tuned models, we take publicly-available pretrained LLMs in their base form and instruction-tune them, both using the default instruction tuning methods and our custom SampleGen method. For default instruction tuning, we follow the method of \citet{wei2022finetuned}, who transform existing datasets into an instructional format. Specifically, they take training examples from 62 publicly available text datasets from TensorFlow Datasets, spanning both language understanding and language generation tasks, and collate them into a single mixture. These datasets are organised into twelve task clusters.
We select a subset of 37 tasks, maintaining the task clusters as depicted in Figure \ref{fig:flan_datasets}. To train both our default FLAN model and our SampleGen model, we require two different training sets: \textit{Regular Prompt} and \textit{SampleGen Prompt}. To create each dataset, we draw from each of the 37 tasks, implementing example-based selection from \citet{wei2022finetuned}. These training splits are detailed in Appendix \ref{app:trainingdetails}. 

\subsection{Models}
We evaluate our methods across a range of model families and scales. Specifically, we use LLaMA-2 and Mistral(7B), and evaluate both the 7B and 13B versions of LLaMA-2. To further ensure robustness, we evaluate across three different random seeds by performing instruction-tuning three separate times. However, due to the prohibitive cost of instruction tuning multiple times, this is done only with the LLaMA-2 13B model. We detail our full training hyperparameters in Appendix \ref{app:subsec:hyperparams}.

At the heart of the discussion on the capabilities of LLMs is the question of whether some (larger) models exhibit behaviour that is `qualitatively different'~\cite{wei2022emergent}. Without taking a stance on whether this claim holds, our choice of the LLaMA-2 13B model is further motivated by prior research demonstrating its ability to exhibit such behaviour~\cite{lu-etal-2024-emergent}. Therefore, the core of our experiments focuses on the LLaMA-2 13B model, supplemented by additional experiments on other models to demonstrate generalisability across different scales and model families.

\section{Correlating the Performance of Base Models and Instruction-Tuned Models}
\label{sec:rq1}
This section describes our experiments aimed at answering the research question presented in Section \ref{sec:intro}. Specifically, we evaluate the relationship between the performance of base LLMs and instruction-tuned LLMs. This comparison is performed by evaluating the correlation between the performance of the five model settings described earlier in Table \ref{tab:evaltypes}: Base Direct, Gold-Pipeline, FLAN, SampleGen, and SampleGen-Pipeline. 

\subsection{Evaluation Tasks and Metrics}
\label{sec:evaluation_tasks_and_metrics}
We select four NLU clusters and two NLG clusters for evaluation. From each cluster, we choose one task included in training and one that is not. For the NLI cluster specifically, we select an additional out-of-training task to evaluate within-cluster consistency. Additionally, we include a single NLG task from a cluster entirely excluded from training, providing a completely out-of-distribution evaluation. This setup is illustrated in Figure \ref{fig:flan_datasets}.

As previously discussed, a key aspect of LLM capabilities is the \emph{emergence} of certain abilities, demonstrated by their capacity to perform above the random baseline on specific tasks without being explicitly trained on them. To evaluate the relationship between these capabilities in base LLMs and instruction-tuned LLMs, we include in our evaluation three tasks previously identified as `emergent' and three tasks not classified as `emergent'~\cite{wei2022emergent}. These tasks are a subset of the BigBench dataset~\cite{srivastava2023beyond}. It is important to note that this prior identification of `emergence' was based on the performance of GPT models, so our results may vary slightly. However, our goal is to evaluate whether models across the five different settings we test in consistently succeed or fail at solving these tasks.

For evaluation, we employ the default evaluation metrics of each task, as shown in Table \ref{app:tab:original_metrics} in Appendix \ref{app:task_metrics}. However, for tasks that are evaluated by accuracy, we instead employ the more flexible BERTScore Accuracy, following \citet{lu-etal-2024-emergent}. This ensures that answers that do not match the target exactly, but are nevertheless correct, can be identified as such.

\subsection{Results}

Figure \ref{fig:main_charts} provides an illustration of our results. Each spoke in the radar chart represents a task and each of the plots represents the performance of LLaMA-2 13B in the different settings described in Table \ref{tab:evaltypes}. The next section presents our results demonstrating generalisability across model scale and families. We also evaluate the Spearman Rank correlation between each pair of settings that we experiment on. Our results demonstrate that there is a \emph{statistically significant} correlation between the performance of LLaMA-2 13B in both the base and instruction-tuned settings. Additionally, the performance of SampleGen-Pipeline demonstrates that instruction tuning allows the model to appropriately generalise to new tasks. More interesting is the fact that the closeness of model performance in the instruction-tuned and base settings is reliant on whether or not the tasks are in-domain with respect to the task clusters. Specifically, we note that the BigBench tasks, which are out-of-domain, are the tasks on which all settings perform most similarly. In particular, we note that models in all our settings are consistent in their ability to perform above the random baseline on the BigBench tasks (which is similar to the performance of Base-Direct, also see Appendix \ref{app:full_results_main_experiments}). In other words, instruction-tuned models are unable to solve any of the tasks that base models cannot solve using in-context learning. We further explore this relation along with other biasing factors in Section \ref{sec:rq2}.

These results demonstrate that the performance of instruction-tuned models is not fundamentally different from that of base models and that there is a statistically-significant correlation between the two. By building on previous work, which has demonstrated that the performance of base models is reliant on the pretraining distribution, we are thus able to show that instruction tuning does not lead to fundamentally novel capabilities, but provides a way to extend base model capabilities based on the specific tasks used in instruction tuning. 

\subsection{Generalisability Across Model Families and Scale}
\label{hyp1_generalizability_subsection}

To ensure generalisability, we evaluate multiple models across a range of families and scales, following the approach described earlier. Figures \ref{app:fig:llama_7B_hyp1} and \ref{app:fig:mistral_7B_hyp1} in the Appendix present the spider plots and correlation statistics for these models, while detailed numerical results are provided in Appendix \ref{app:full_results_main_experiments}. Overall, our findings indicate that, across model families and scales, the performance of instruction-tuned models is closely correlated with that of base models.

\section{Evaluating Possible Confounding Factors}
\label{sec:rq2}
Having established the correlation between the capabilities of base and instruction-tuned models, we focus on exploring how various confounding factors might impact this correlation.  These experiments aim to strengthen our findings.

\subsection{Prompt Complexity}
\label{sec:main:prompt_complexity}
First, we evaluate whether the correlation between the performance of base and instruction-tuned models holds across varying levels of prompt complexity. Specifically, we investigate whether increasing the complexity of the prompt has a similar impact on the performance of both types of models. If the performance of instruction-tuned models does not degrade in a manner similar to that of base models, it may suggest that instruction-tuned models operate in a fundamentally different way.

We detail the prompts used in these experiments, and the method of generating these prompts, in Appendix \ref{app:full_results_prompt_complexity}. The results of our experiments, which confirm the continued correlation between the performance of base models using in-context learning and instruction-tuned models, are presented in Figure \ref{fig:prompt-effectiveness}.

Since the notion of prompt complexity is contentious, we add an additional setup where we translate the prompt to a low-resource language, namely Scottish Gaelic. This provides an assurance that we have a more difficult prompt than the standard (English) prompt. In this setting as well, we observe that both base and instruction-tuned models perform poorly, confirming our previously-observed correlation. We likewise detail the method for generating these prompts, as well as the full results, in Appendix \ref{app:full_results_prompt_complexity}. 

\subsection{Semantic Data Labels} 
We also experiment with replacing the semantic tags associated with the components of a task sample with non-semantic labels. For example, we replace ``Premise'' and ``Hypothesis'' in the task description of an NLI sample with ``Input 1'' and ``Input 2''. If an instruction-tuned LLM has better generalisation capabilities than a base model, then we expect the performance of instruction-tuned LLMs on the semantically-neutral task samples to be higher than that of the base model. 

Our experiments, the results for which are presented in Appendix \ref{app:unified_vs_non}, showed that replacing semantic data labels with generic ones in fact leads to deteriorating performance in both the base and the instruction-tuned models. This further demonstrates that LLMs are likely extrapolating from pretraining data, as semantic labels provide a more efficient way of doing so.

\subsection{Instruction-Tuning Tasks}
\label{subsec:instruction-tuning-tasks}
Finally, we test whether the specific tasks included in the instruction-tuning data influence the generalisation of instruction-tuned models.

\begin{figure}
    \centering
    \includegraphics[width=0.9\linewidth,trim={1.5cm 0cm 1cm 7cm},clip]{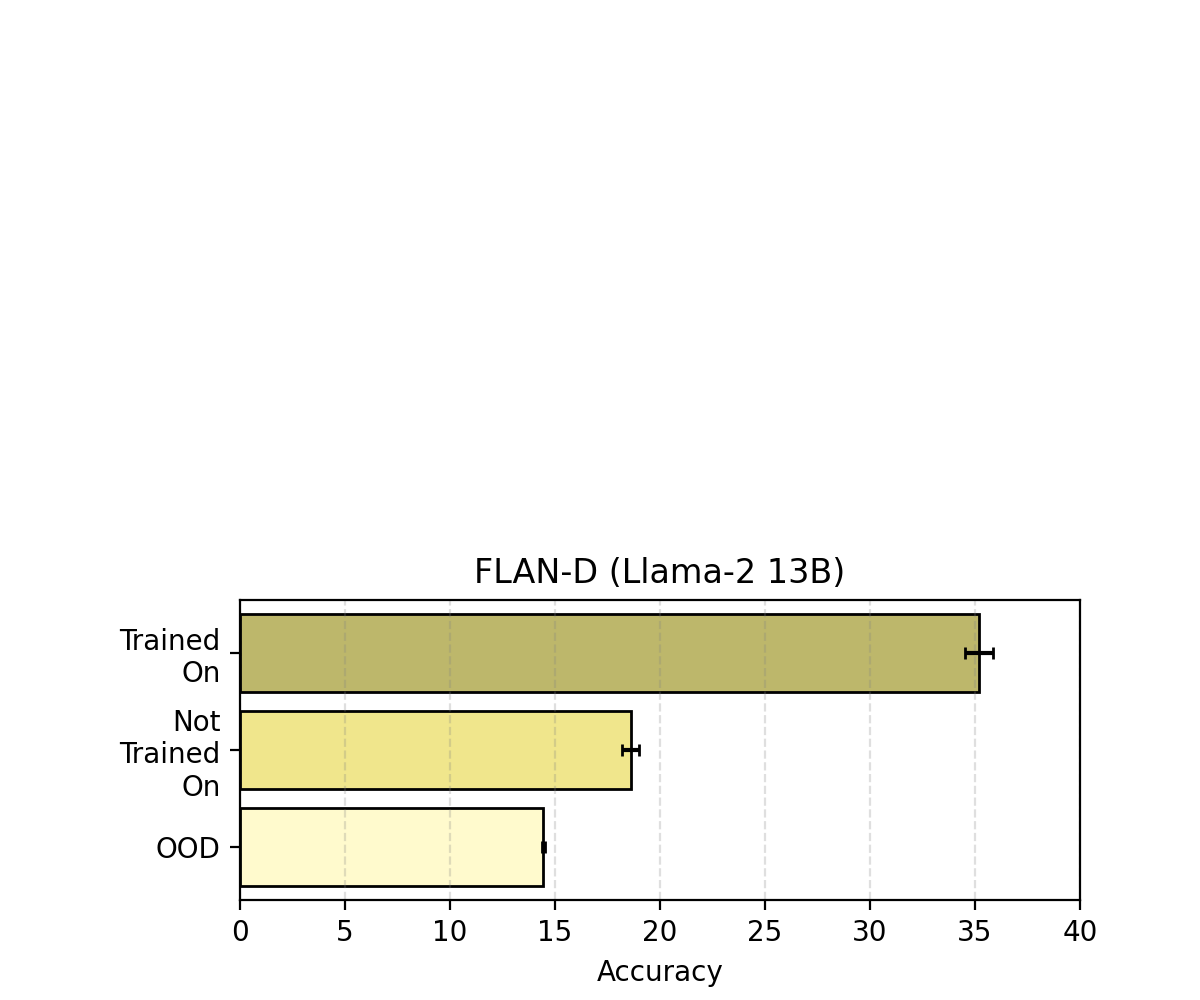}
    \caption{Overall performance trend of our Llama-2 13B FLAN model across three clusters of tasks, with standard deviation across 3 seeds. The x-axis is truncated to terminate at 40.
    }
    \label{fig:normalized_bars_main_experiments}
\end{figure}

We conduct our analysis on our \textsc{FLAN} model. First, we aggregate our experiment results: We take 4 tasks that were used to instruction-tune \textsc{FLAN}: MNLI, SNLI, BoolQ, and Winogrande; 3 \textit{related tasks}: QNLI, StoryCloze, WSC273; and 6 \textit{out-of-distribution tasks}: SocialIQA, Logical Deduction, Strange Stories, Tracking Shuffled Objects, GSM8K, and Causal Judgment. To obtain a single score for each of these three groups, we subtract the random baselines of each task from the raw task scores, then average over the number of tasks in the group. The resulting scores in Figure \ref{fig:normalized_bars_main_experiments} confirm our expectations, showing that \textsc{FLAN} is more adept at related tasks, less so to unrelated tasks, and finally the least adept at tasks that are out-of-distribution.

To confirm this trend more rigorously, we conduct an additional set of experiments, in which we control for the specific tasks used in instruction-tuning. Given a target task, we define the notion of \textit{related} and \textit{unrelated} tasks by means of the 12 task clusters and 2 task categories shown in Figure \ref{fig:flan_datasets}. According to this specification, a task that is related to the target task will be any task that belongs to the \textit{same cluster} (e.g. NLI), while an unrelated task will be a task that belongs to one of the following: 1) A \textit{different cluster}, but still within the \textit{same category} as the target task (e.g., NLU), 2) A \textit{different category} than the target task (NLU vs. NLG). Finally, a task that belongs to \textit{none of the clusters} within either of the two categories (i.e., the BigBench tasks and GSM8K) is considered \textit{out-of-distribution}. Importantly, we do not base our classification on semantics- or distance-based similarity between the task datasets themselves.

For our experiments, we select two tasks from the NLI cluster shown in Figure \ref{fig:flan_datasets} for evaluation: QNLI and MNLI-Matched. We instruction-tune four different models as follows: \textbf{1) Target Task:} one model on the target task and all other clusters except for NLI; \textbf{2) Adjacent Task:} one model on an NLI task related to the target and all other clusters except for NLI; \textbf{3) Adjacent Category:} one model on all NLG and NLU clusters except for NLI; \textbf{4) NLG:} one model exclusively on the NLG tasks. 
This approach allows us to carefully control for the tasks used in evaluation and training, and thereby better assess their impact.
 Table \ref{tab:hyp2evaltypes} in the Appendix summarizes the four models used in these experiments, along with the particular tasks used in their instruction-tuning.

\begin{figure}
    \centering
    \includegraphics[width=0.95\linewidth,trim={3cm 1cm 9.5cm 14.5cm},clip]{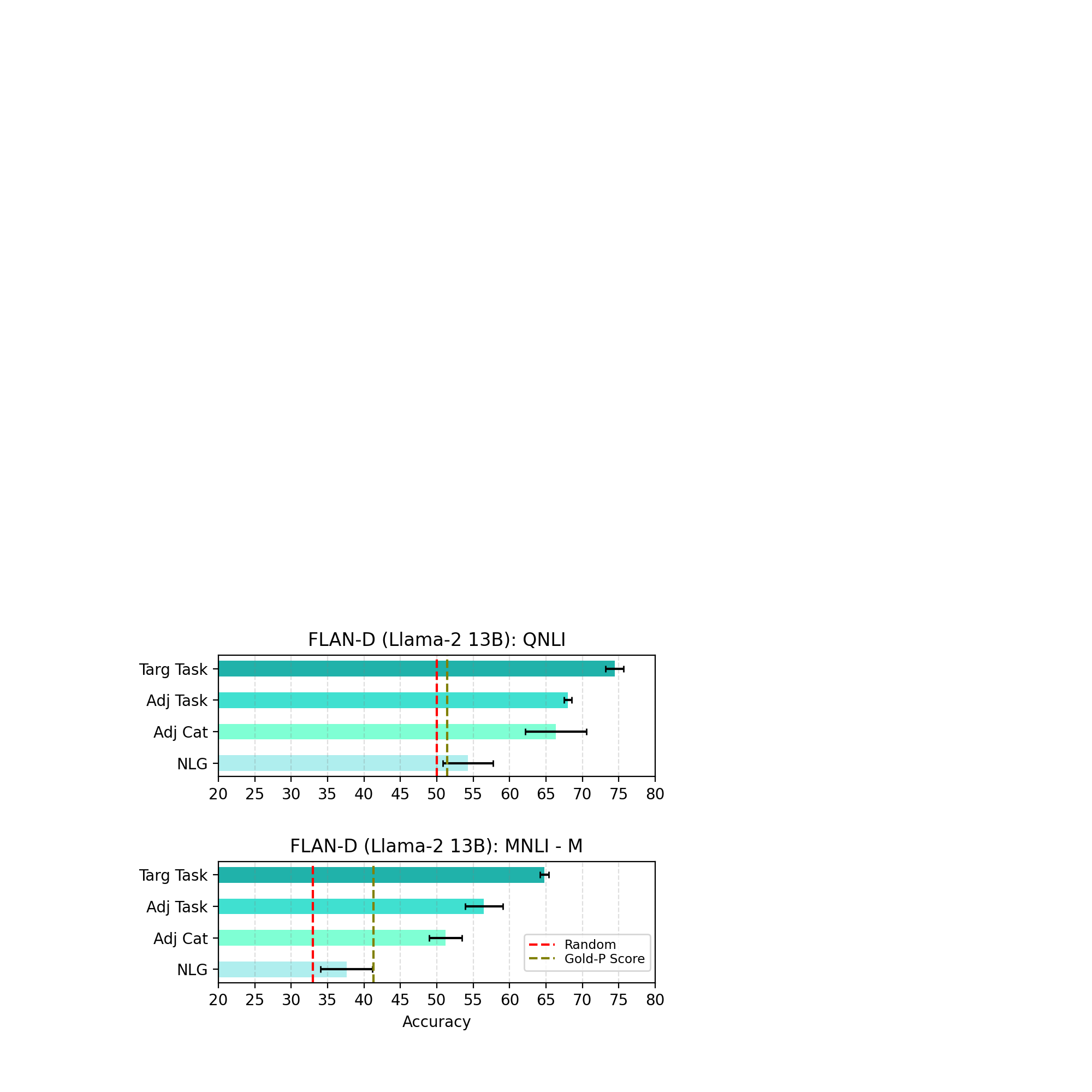}
    \caption{Overall performance trend of our Llama-2 13B FLAN model on QNLI and MNLI-Matched, with standard deviation over 3 seeds. The x-axis is truncated to begin at 20 and terminates at 80.
    }
    \label{fig:hyp2_nli}
\end{figure}

These results, illustrated in Figure \ref{fig:hyp2_nli}, demonstrate that the `closeness' of a target task to the tasks in the instruction-tuning data has a consistent impact on the performance of a model on that task. 

Although this trend is not statistically significant, we hypothesize that this is due to the small number of target tasks examined. We additionally show that our results hold across various model scales, as presented in Appendix \ref{app:task_ablation_experiments}.

\section{Conclusion and Implications}

Our work sheds light on whether or not the capabilities of LLMs extend beyond extrapolation from pretraining data. By demonstrating a significant correlation between the abilities of instruction-tuned models and their base counterparts, we can leverage prior research on base models to show, by extension, that a relationship exists between an instruction tuned LLM's ability to respond to a query and the possibility to extrapolate priors from pretraining data to answer that query, with the added factor of instruction tuning data.

A key contribution of this work is its ability to explain the seemingly contradictory capabilities of LLMs. The reliance on pretraining data helps clarify why these models can excel at solving complex college-level tasks while failing at simpler ones that human children can handle. Similarly, while some studies suggest that LLMs exhibit sparks of intelligence~\cite{bubeck2023sparks} or embers of autoregression~\cite{doi:10.1073/pnas.2322420121}, we argue that these capabilities are better understood as the ability to generalise from pretraining data.

The possibility that LLMs can solve tasks well beyond their training data has been used to claim that they pose an existential threat~\cite{wei2022emergent}. Our findings, like other prior work~\cite{lu-etal-2024-emergent}, contradict this position, suggesting that such concerns are overstated. While extrapolation from pretraining data---particularly when such data includes all of the web---is undeniably powerful, it remains limited to extending existing patterns rather than enabling truly novel inference steps. The bounds we establish have significant implications, as understanding how LLMs respond to prompts and what their performance depends on carries real-world consequences that warrant careful evaluation.

\section{Limitations}
While knowledge of a model's exact pretraining data is not necessary for our argument, there is a possibility that some of the evaluation tasks we used could have been seen by our models (namely, Llama-2 and Mistral) during pretraining. This, in turn, could possibly affect the models' downstream performance on these tasks. Moreover, since the pretraining datasets of these models is not publicly known, this possibility cannot be immediately excluded. Conducting a data contamination analysis is beyond the scope of our work.

In discussing the similarity of tasks, we rely on human-defined distinctions that are well-established in NLP (e.g. NLI tasks, Coreference tasks, summarization tasks). We deliberately do not conduct a semantic- or distance-based similarity analysis between tasks to determine our task clusters, as we consider this aimed towards a different research question than the one we pose. However, we acknowledge that employing such a clustering method might identify different similarity relations between tasks than our present categorization, and might be an interesting exploration for subsequent work.

\section*{Acknowledgements}
This work was funded by the LOEWE Distinguished Chair “Ubiquitous Knowledge Processing”, LOEWE initiative, Hesse, Germany (Grant Number: LOEWE/4a//519/05/00.002(0002)/81), as well as by the German Federal Ministry of Education and Research and the Hessian Ministry of Higher Education, Research, Science and the Arts within their joint support of the National Research Center for Applied Cybersecurity ATHENE.

\bibliography{acl_latex}
\newpage

\appendix

\section{In-Context Learning}
\label{app:icl}
In this section, we provide further details pertaining to in-context learning. 

~\citet{wies2023the} show that ICL might be more about identifying the task rather than learning it, and both of these imply the reliance on pretraining data. However, others have put forward the `structured task hypothesis', which indicates that ICL might well extend beyond the idea of selecting tasks and is capable of solving tasks that are not present in pretraining data~\cite{li-etal-2024-language,wei2024larger}. This generalisation capability is addressed by different hypotheses of how LLMs perform in-context learning which suggests that in-context learning can be thought of as an implicit form of Bayesian inference. ~\citet{xie2022an} and \citet{chen-etal-2024-parallel} suggest that parallel structures in pretraining data lead to in-context capabilities. Other work in this area includes the use of mechanistic interpretability~\cite{cho2024revisitingincontextlearninginference}.

While the exact mechanism for how LLMs perform ICL is still under exploration, what all these works have in common is that ICL is a form of extrapolation from pretraining data. Indeed, some work has explicitly shown that the frequency of an item in pretraining data has a direct impact on the ability of a model to reason about it~\cite{razeghi-etal-2022-impact}. Of particular relevance to our work is the finding that there is no evidence to suggest that base models can do more than extrapolate from their pretraining data~\cite{zhou-etal-2024-mystery}. This suggests that base LLMs respond to questions using ICL by making use of the examples provided to isolate and extrapolate relevant priors from their pretraining data.
\section{Training Details}
In this section, we present the full details of our training setup.

\subsection{Hyperparameters and Training Time}
\label{app:subsec:hyperparams}
Our experiments involved instruction-tuning 90 models across various setups. We use two kinds of instructional prompts: \textit{Regular Prompt} and \textit{SampleGen Prompt}. Table \ref{app:tab:training_hyperparams} presents the training details of each kind of model. We train our models on the following GPU architectures: NVIDIA TITAN RTX, NVIDIA RTX A6000, NVIDIA A100 80GB PCIe. Inference was performed on L40 48GB GPUs. We use the following architectures: Mistral 7B, Llama-2 7B, and Llama-2 13B.

We note that a potential risk of instruction-tuning many models is the inherent environmental cost of computation. For this reason, we instruction-tuned all models on a low number of epochs. Additionally, to keep the training and inference times manageable, we restricted the number of in-context samples in SampleGen Prompt to 2 and did not experiment with higher values. 

\begin{table*}[htp]
\footnotesize
\centering
\begin{tabular}{rccccc}
\toprule
\textbf{Prompt Type} & \textbf{Num Train}& \textbf{Num Epochs}  & \textbf{Batch Size} & \textbf{Average Training Time} & \textbf{Num GPUs}\\ \midrule
Regular Prompt & 20K & 1 & 2 & 8 hrs & 1\\
SampleGen Prompt & 20K & 1 & 2 & 14 hrs & 1\\
\bottomrule
\end{tabular}
\caption{\label{app:tab:training_hyperparams}Training details of our FLAN (Regular Prompt) and SampleGen models.}
\end{table*}

\subsection{Training Data}
\label{app:trainingdetails}

Table \ref{tab:training_data_stats} shows the details of the datasets we use for instruction-tuning our models. The language of all tasks in these datasets is English, with the exception of the tasks in the \enquote{Translation} category, for which the languages are English and the target language for translation (detailed in Figure \ref{fig:flan_datasets}).

We create the Regular Prompt dataset by picking 37 of the 62 tasks in \citet{wei2022finetuned}, as described in Section \ref{sec:instruction-tuning-and-evaluation-datasets}. To the best of our knowledge, at the time of writing there was no publicly-available version of the original FLAN dataset of \citet{wei2022finetuned} that was compatible with HuggingFace\footnote{\url{https://huggingface.co/}}, our chosen framework for our experiments. For this reason, we reimplemented the data loading for all tasks found in the original FLAN dataset, using the publicly-available code of \citet{wei2022finetuned} as a basis.\footnote{\url{https://github.com/google-research/FLAN/blob/main/flan}} The code of \citet{wei2022finetuned} was distributed under the Apache License, Version 2.0, and we adhere to the terms of the license in our repository, e.g. by designating our code as a derivative and retaining a link to the source.

To create our final training dataset, we implement example-based mixing \cite{wei2022finetuned}, which proportionally draws samples from each dataset according to the dataset sizes. The resulting training mixture contains 63,556 samples, however we randomly select 20,000 for training our \textsc{FLAN} model.

We create the SampleGen Prompt dataset by partitioning the training split of each task into two parts. From the first part, we pick two samples to serve as in-context samples, and from the second part, we take one sample to serve as the \enquote{main sample}, e.g. the question to be answered. Next, we similarly use example-based mixing to create a training dataset that contains samples from each task, proportional to the original dataset sizes. The resulting training set contains 74,057 examples, from which we pick 20K for training \textsc{SampleGen}.

\section{Full Results}
\label{app:full_results}
In this section, we present the full numerical results of our experiments. In our evaluations, we used both datasets from our re-implemented FLAN dataset, as well as datasets from other sources: BIG-Bench and GSM8K. The BIG-Bench dataset is licensed under the Apache License, Version 2.0\footnote{\url{https://github.com/google/BIG-bench/tree/main?tab=Apache-2.0-1-ov-file}}, and our use of the dataset, which does not include any modifications, is consistent with the license terms. The GSM8K dataset, which was only used for testing, is licensed under the MIT License\footnote{\url{https://github.com/openai/grade-school-math?tab=License-1-ov-file}}, and our usage is likewise consistent with the license terms.

\subsection{Task Metrics}
\label{app:task_metrics}
Table \ref{app:tab:original_metrics} presents the evaluation tasks, along with the default metrics.

We use the HuggingFace implementation\footnote{\url{https://huggingface.co/spaces/evaluate-metric/bertscore}} of BERTScore \cite{zhang2020bertscoreevaluatingtextgeneration}, and use the hyperparameters from \citet{lu-etal-2024-emergent}. For our statistical significance tests, we use the Scipy implementation \textit{spearmanr}\footnote{\url{https://docs.scipy.org/doc/scipy/reference/generated/scipy.stats.spearmanr.html}} using default parameters. For other metrics, e.g. ROUGE1 and BLEU, we use the same implementations \footnote{\url{https://github.com/google-research/FLAN/blob/main/flan/metrics.py}} used in \citet{wei2022finetuned} without modification, which were distributed under the Apache License, Version 2.0.

\begin{table}
\footnotesize
    \centering
    \begin{tabular}{rc}
    \toprule
\textbf{Task}  &  \textbf{Original Metric}  \\
\midrule
NLI    \\ 
\hline
QNLI  &  Accuracy  \\
MNLI M  &  Accuracy  \\
MNLI MM  &  Accuracy  \\
SNLI  &  Accuracy  \\
StoryCloze  &  Accuracy  \\
CoPA  &  Accuracy  \\
SQuAD  &  F1  \\
BoolQ  &  Accuracy  \\
Winogrande  &  Accuracy  \\
WSC273 (OD)  &  Accuracy  \\
\hline
NLG     \\
\hline
DART  &  Rouge1  \\
CommonGen  &  Rouge1  \\
WMT-16 EN/DE  &  BLEU  \\
WMT-16 EN/RO  &  BLEU  \\
Gigaword  &  Rouge1  \\
\hline
BB - Emergent   \\
\hline
SocialIQA  &  Accuracy  \\
LogicalDeduction  &  Accuracy  \\
Strange Stories  &  Accuracy  \\
\hline
BB - Not Emergent    \\
\hline
Tracking (7)  &  Accuracy  \\
GSM8K  &  Accuracy  \\
Causal judgement  &  Accuracy  \\
\bottomrule
    \end{tabular}
    \caption{The evaluation tasks used in our main experiments, along with the original metrics for each task. For tasks with Accuracy as their original metric, we instead use BERTScore Accuracy, as detailed in Section \ref{sec:evaluation_tasks_and_metrics}}
    \label{app:tab:original_metrics}
\end{table}

\begin{table}
\footnotesize
\begin{center}
\begin{tabular}{rcc}
\toprule
\textbf{Model Comparison} & \textbf{Spearman r} & \textbf{p-value}  \\
\midrule
FLAN - SG Direct & 0.776 & 2.99E-03 \\
FLAN - SG Pipeline & 0.713	& 0.0092\\
SG Direct - SG Pipeline & 0.357	& 0.2551 \\
\bottomrule
\end{tabular}
\end{center}
\caption{\label{tab:hyp2_statistics} Correlation and statistical significance scores for our NLI experiments demonstrating correlation. We note that the correlation isn't statistically significant and hypothesize that this is due to the fact that we test a small set of tasks.
}
\end{table}

\subsection{Main Experiments}
\label{app:full_results_main_experiments}
This section presents the results for our main experiments. In Table \ref{app:hyp1_llama13B_avg}, we present the results for Llama-2 13B averaged across our 3 seeds, and the results for the individual seeds in Tables \ref{app:hyp1_llama13B_seed1} \ref{app:hyp1_llama13B_seed2}, and \ref{app:hyp1_llama13B_seed3}. Results for Llama 7B are presented in Table \ref{app:hyp1_llama7B}, and results for Mistral 7B are presented in Table \ref{app:hyp1_mistral7B}.

\begin{table}[htp]
\footnotesize
\centering
\begin{tabular}{p{2cm}p{2cm}p{2cm}}
\toprule
\textbf{Dataset} & \textbf{Num Samples}& \textbf{Used for Training} \\ \midrule
Regular Prompt & 63,556 & 20,000 \\
SampleGen Prompt & 74,057 & 20,000 \\
\bottomrule
\end{tabular}
\caption{\label{tab:training_data_stats}Details of our training datasets for the main experiments.}
\end{table}

\begin{figure*}
    \centering
    \stackunder[0pt]{\includegraphics[width=0.8\linewidth,trim={1cm 5cm 0cm 7cm},clip]{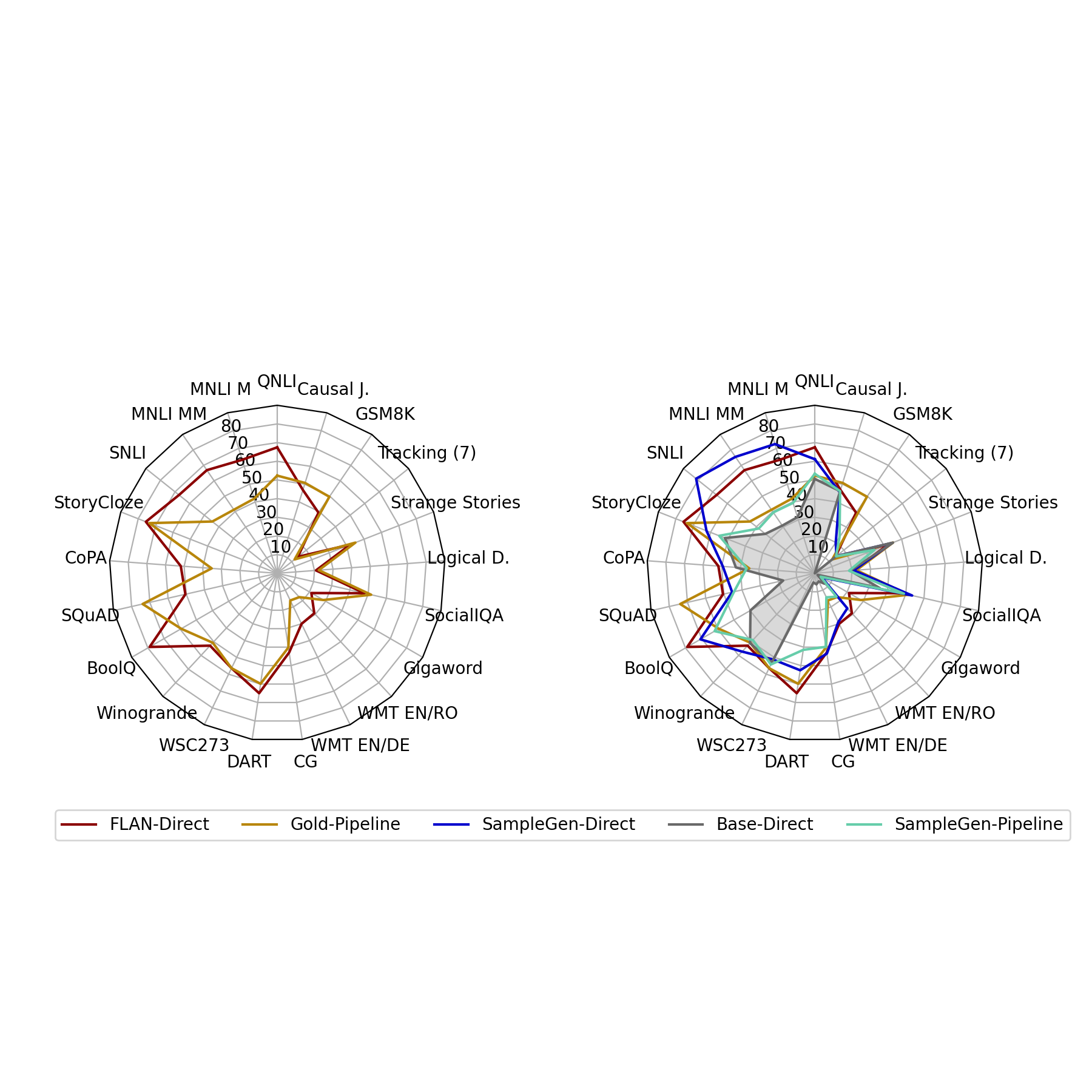}}{}%
    \stackunder[0pt]{\includegraphics[width=0.5\linewidth,trim={0cm 7cm 15cm 0cm},clip]{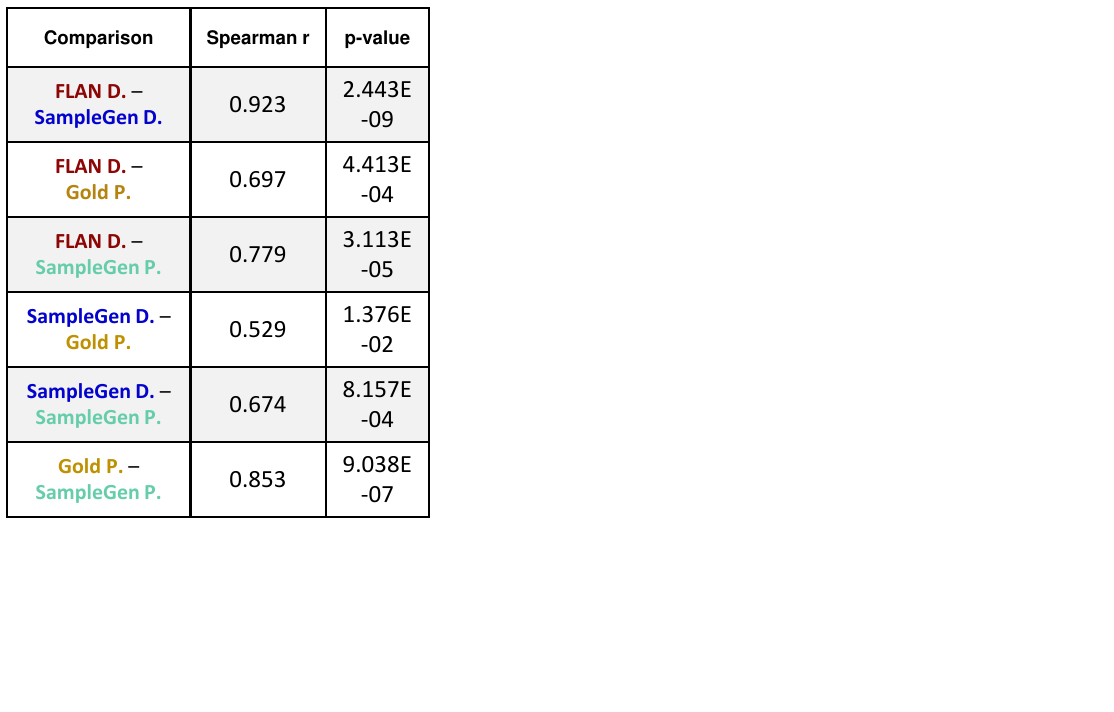}}{}
    \caption{Correlation and statistical test results for our main experiments on Llama-2 7B, showing the significant correlation between the performance of base models prompted with in-context examples (and their variants) and that of instruction-tuned models.}
    \label{app:fig:llama_7B_hyp1}
\end{figure*}

\begin{figure*}
    \centering
    \stackunder[0pt]{\includegraphics[width=0.8\linewidth,trim={1cm 5cm 0cm 7cm},clip]{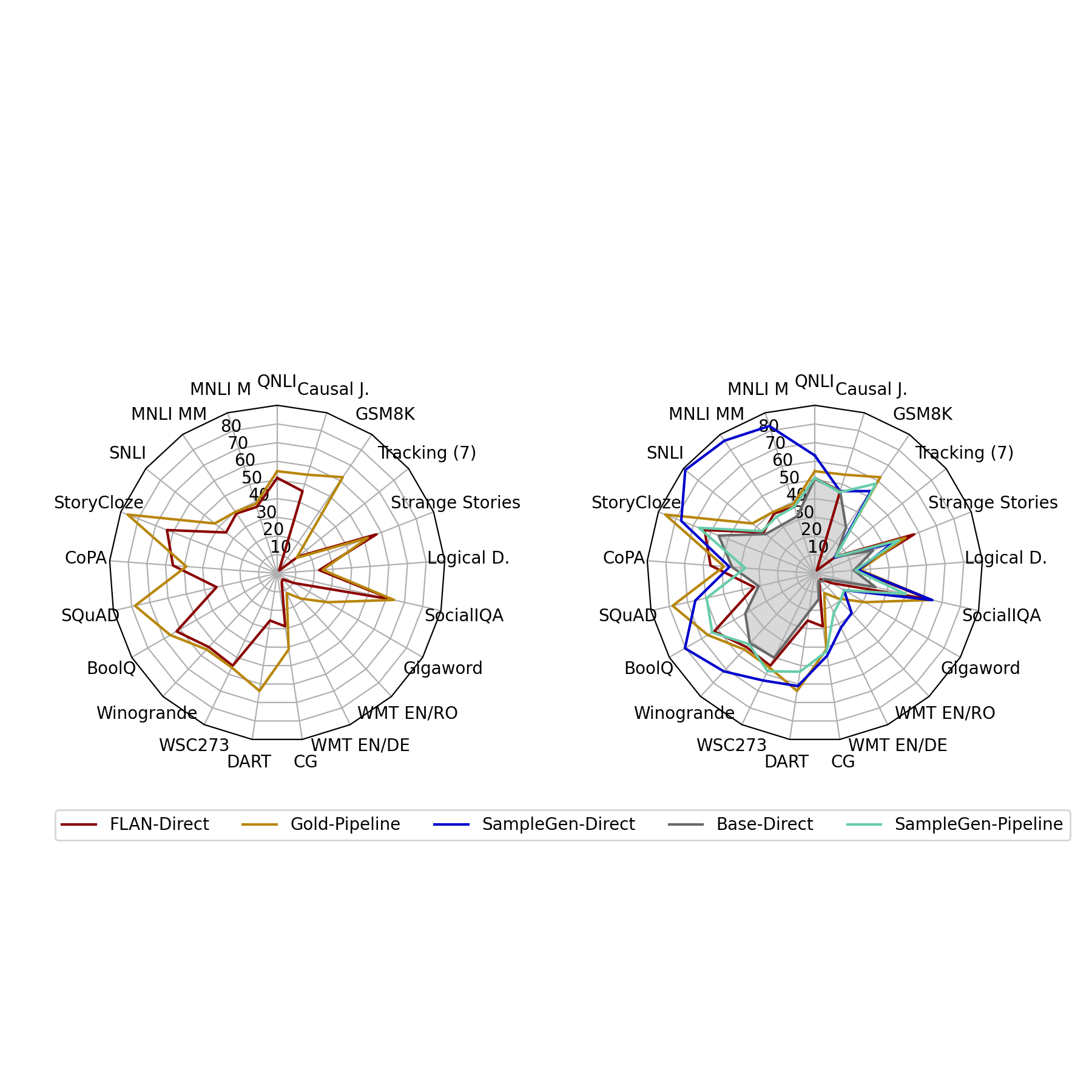}}{}%
    \stackunder[0pt]{\includegraphics[width=0.5\linewidth,trim={0cm 7cm 15cm 0cm},clip]{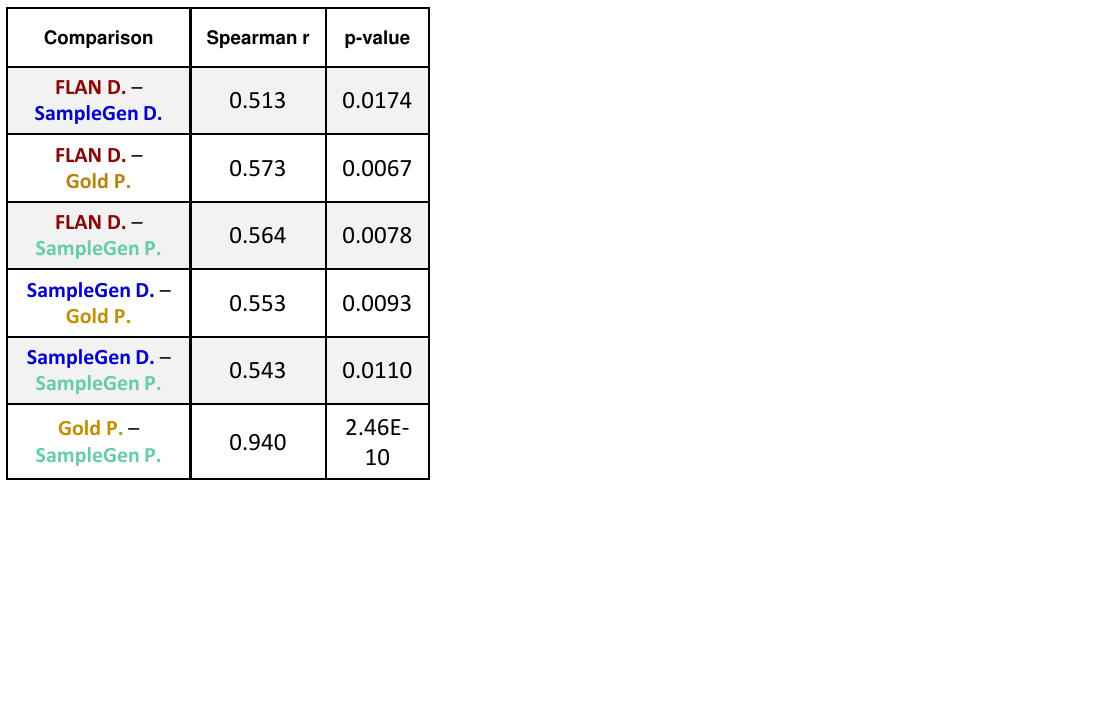}}{}
    \caption{Correlation and statistical test results for our main experiments on Mistral 7B, showing the significant correlation between the performance of base models prompted with in-context examples (and their variants) and that of instruction-tuned models.}
    \label{app:fig:mistral_7B_hyp1}
\end{figure*}

\begin{figure*}
    \centering
    {\includegraphics[width=0.8\linewidth,trim={0cm 5cm 0cm 7cm},clip]{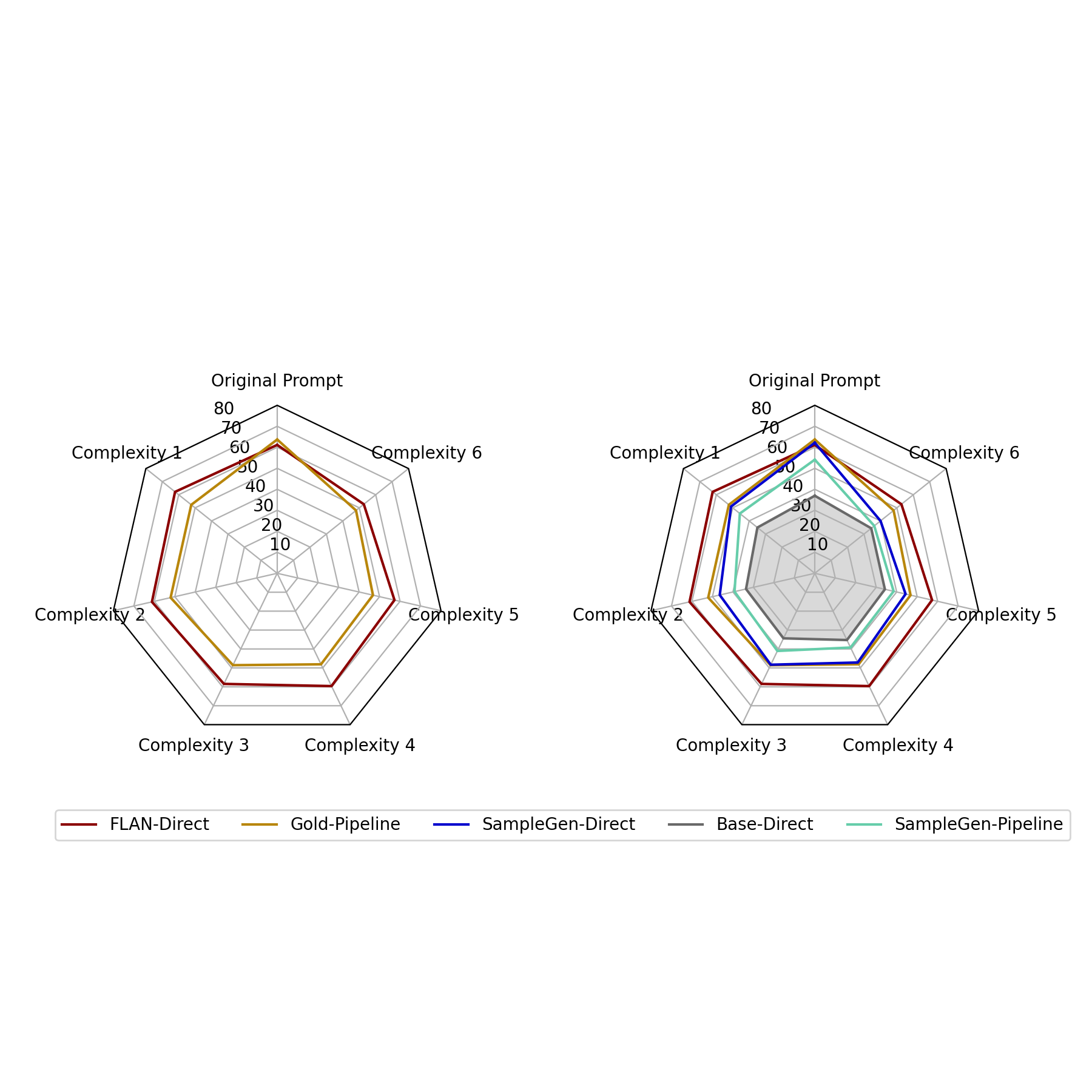}}{}%
    \caption{The performance of our LLaMA-2 13B models on SocialIQA across prompts of different complexity, demonstrating the continued correlation between base models tested using in-context examples and instruction-tuned models across prompts of different complexity.
    }
    \label{fig:prompt-effectiveness}
\end{figure*}

\begin{table*}[!htb]
\footnotesize
\centering
\begin{tabular}{rccccccc}
\hline
\textbf{Test Task}                        & \textbf{Metric} & \textbf{Random} & \textbf{Base (D)} & \textbf{FLAN (D)} & \textbf{SampleGen (D)} & \textbf{Gold (P)} & \textbf{SampleGen (P)} \\
\hline
NLU \\
\hline
QNLI & BSA & 50 & 51.54 & 61.73 & 60.68 & 51.27 & 48.43 \\
MNLI M & BSA & 33  & 32.12 & 73.92 & 79.48 & 41.53 & 37.69 \\
MNLI MM & BSA & 33 & 31.32 & 74.93 & 82.61 & 42.89 & 37.08 \\
SNLI & BSA & 33  & 33.88 & 80.73 & 87.33 & 37.04 & 35.88 \\
StoryCloze & BSA & 50  & 50.81 & 81.55 & 79.21 & 82.11 & 63.23 \\
CoPA & BSA & 50  & 34.20 & 48.60 & 45.47 & 37.80 & 25.00 \\
SQuAD & F1 & 50 & 9.12 & 70.39 & 55.10 & 78.96 & 54.49 \\
BoolQ & BSA & 50  & 41.84 & 82.30 & 76.47 & 65.13 & 65.11 \\
Winogrande & BSA & 50 & 48.38 & 63.01 & 66.77 & 53.54 & 51.41 \\
WSC273 (OD) & BSA & 50 & 49.37 & 62.51 & 61.66 & 58.12 & 57.75 \\
\hline
NLG \\
\hline
DART & Rouge1& --  & 3.32 & 67.42 & 15.81 & 62.39 & 13.22 \\
CommonGen & Rouge1 & -- & 7.15 & 43.34 & 43.44 & 41.84 & 39.20 \\
WMT-16 EN/DE & BLEU & --  & 2.74 & 33.16 & 33.43 & 12.63 & 25.46 \\
WMT-16 EN/RO & BLEU  & -- & 5.26 & 32.26 & 29.78 & 19.19 & 22.21 \\
Gigaword (OD) & Rouge1 & --  & 0.36 & 21.28 & 13.03 & 31.26 & 13.48 \\
\hline
BB - Emergent  \\
\hline
SocialIQA & BSA & 33  & 36.93 & 61.13 & 62.18 & 63.71 & 54.11 \\
LogicalDeduction & BSA & 20  & 18.86 & 26.64 & 22.17 & 28.70 & 21.19 \\
Strange Stories & BSA & 50  & 37.62 & 50.24 & 45.48 & 52.62 & 39.76 \\
\hline
BB - Not Emergent  \\
\hline
Tracking (7) & BSA & 14 & 14.62 & 13.19 & 15.12 & 14.00 & 15.17 \\
GSM8K & EMA & 25 & 0.68 & 50.98 & 50.22 & 58.21 & 52.01 \\
Causal judgement& BSA & 50  & 46.93 & 46.49 & 48.03 & 50.22 & 46.49 \\
\hline
\end{tabular}
\caption{Full main experiment results for Llama-2 13B models, averaged across 3 seeds.}
\label{app:hyp1_llama13B_avg}
\end{table*}

\begin{table*}[htb]
\footnotesize
\centering
\begin{tabular}{rccccccc}
\hline
\textbf{Test Task}                        & \textbf{Metric} & \textbf{Random} & \textbf{Base (D)} & \textbf{FLAN (D)} & \textbf{SampleGen (D)} & \textbf{Gold (P)} & \textbf{SampleGen (P)} \\
\hline
NLU &  &  &  &  &  &  \\
\hline
QNLI & BSA & 50 & 53.50 & 60.06 & 66.04 & 50.72 & 49.16  \\
MNLI M & BSA & 33 & 32.12 & 74.32 & 79.96 & 41.68 & 38.16  \\
MNLI MM & BSA & 33  & 31.32 & 75.64 & 83.44 & 43.12 & 37.64  \\
SNLI & BSA & 50  & 33.88 & 81.04 & 87.64 & 37.44 & 36.72  \\
StoryCloze & BSA & 50  & 51.42 & 81.03 & 76.32 & 81.29 & 61.79  \\
CoPA & BSA & 50  & 34.20 & 46.40 & 45.00 & 37.40 & 20.60  \\
SQuAD & F1 & 50 & 10.77 & 60.82 & 43.12 & 78.12 & 42.44  \\
BoolQ & BSA & 50  & 41.84 & 82.57 & 77.36 & 65.16 & 65.84  \\
Winogrande & BSA  & 50 & 48.38 & 63.14 & 67.88 & 53.75 & 51.38  \\
WSC273 (OD) & BSA & 50  & 47.02 & 63.00 & 62.64 & 56.41 & 60.44  \\
\hline
NLG &  &  &  &  &  &   \\
\hline
DART & Rouge1 & --  & 3.32 & 67.37 & 15.71 & 62.50 & 12.40  \\
CommonGen & Rouge1 & -- & 7.15 & 43.42 & 42.42 & 41.45 & 38.87  \\
WMT-16 EN/DE & BLEU & --  & 2.74 & 32.98 & 33.63 & 12.73 & 25.29  \\
WMT-16 EN/RO  & BLEU & -- & 5.26 & 32.39 & 27.54 & 19.02 & 18.36  \\
Gigaword (OD) & Rouge1 & -- & 0.36 & 21.43 & 11.94 & 31.49 & 12.32  \\
\hline
BB - Emergent &  &  &  &  &  &   \\
\hline
SocialIQA & BSA & 33 & 33.40 & 60.92 & 65.37 & 62.98 & 57.30  \\
LogicalDeduction& BSA & 20  & 18.42 & 27.75 & 22.25 & 28.42 & 21.08  \\
Strange Stories  & BSA & 50& 37.14 & 50.00 & 47.86 & 48.57 & 41.43  \\
\hline
BB - Not Emergent &  &  &  &  &  &   \\
\hline
Tracking (7) & BSA & 14  & 14.86 & 13.21 & 15.21 & 13.86 & 15.36  \\
GSM8K & EMA & 25  & 0.68 & 51.03 & 49.44 & 57.85 & 53.37  \\
Causal Judgement& BSA & 50  & 48.68 & 46.05 & 49.34 & 49.34 & 47.37  \\
\hline
\end{tabular}
\caption{Full main experiment results for Llama-2 13B models, seed 42.}
\label{app:hyp1_llama13B_seed1}
\end{table*}

\begin{table*}[htb]
\footnotesize
\centering
\begin{tabular}{rccccccc}
\hline
\textbf{Test Task}                        & \textbf{Metric} & \textbf{Random} & \textbf{Base (D)} & \textbf{FLAN (D)} & \textbf{SampleGen (D)} & \textbf{Gold (P)} & \textbf{SampleGen (P)} \\
\hline
NLU &  &  &  &  &  &  \\
\hline
QNLI  &  BSA  & 50 &  50.56  &  64.64  &  60.04  &  51.68  &  47.80  \\
MNLI M  &  BSA  & 33 &  32.12  &  74.88  &  78.68  &  41.60  &  37.60  \\
MNLI MM  &  BSA  & 33 &  31.32  &  75.52  &  82.24  &  43.20  &  36.64  \\
SNLI  &  BSA  & 50 &  33.88  &  81.88  &  87.28  &  36.68  &  35.52  \\
StoryCloze  &  BSA  & 50 &  50.51  &  79.48  &  79.64  &  82.74  &  63.34  \\
CoPA  &  BSA  & 50 &  34.20  &  50.40  &  42.00  &  37.40  &  24.20  \\
SQuAD  &  F1  & 50 &  8.29  &  75.60  &  59.38  &  79.50  &  59.13  \\
BoolQ  &  BSA  & 50 &  41.84  &  82.12  &  75.92  &  64.68  &  65.76  \\
Winogrande & BSA  & 50  &  48.38  &  63.38  &  65.90  &  53.51  &  51.62  \\
WSC273 (OD) & BSA  & 50  &  50.55  &  62.64  &  60.81  &  60.44  &  56.04  \\
\hline
NLG  &    &    &    &    &    &    \\
\hline
DART  &  Rouge1  & -- &  3.32  &  67.37  &  12.16  &  62.28  &  10.32  \\
CommonGen  &  Rouge1  & --&  7.15  &  43.52  &  44.05  &  42.01  &  38.36  \\
WMT-16 EN/DE  &  BLEU  & --&  2.74  &  33.56  &  33.60  &  12.70  &  26.19  \\
WMT-16 EN/RO   &  BLEU  & --&  5.26  &  32.38  &  31.55  &  19.76  &  24.69  \\
Gigaword (OD)  &  Rouge1  & --&  0.36  &  21.27  &  12.36  &  31.01  &  12.82  \\
\hline
BB - Emergent  &    &    &    &    &    &    \\
\hline
SocialIQA  &  BSA  & 33 &  38.70  &  61.43  &  63.82  &  63.50  &  55.56  \\
LogicalDeduction  &  BSA  & 20 &  19.08  &  28.58  &  22.75  &  27.67  &  20.92  \\
Strange Stories  &  BSA  & 50 &  37.86  &  49.28  &  47.14  &  55.00  &  41.43  \\
\hline
BB - Not Emergent  &    &    &    &    &    &    \\
\hline
Tracking (7)  &  BSA  & 14 &  14.50  &  12.79  &  16.07  &  14.71  &  15.14  \\
GSM8K  &  EMA  & 25 &  0.68  &  50.04  &  52.46  &  58.32  &  49.36  \\
Causal Judgement  &  BSA  & 50 &  46.05  &  46.05  &  46.71  &  51.32  &  45.39  \\
\hline
\end{tabular}
\caption{Full main experiment results for Llama-2 13B models, seed 1234.}
\label{app:hyp1_llama13B_seed2}
\end{table*}

\begin{table*}[htb]
\footnotesize
\centering
\begin{tabular}{rccccccc}
\hline
\textbf{Test Task}                        & \textbf{Metric} & \textbf{Random} & \textbf{Base (D)} & \textbf{FLAN (D)} & \textbf{SampleGen (D)} & \textbf{Gold (P)} & \textbf{SampleGen (P)} \\
\hline
NLU &  &  &  &  &  &  \\
\hline
QNLI  &  BSA & 50  &  50.56  &  60.48  &  55.96  &  51.40  &  48.32  \\
MNLI M  &  BSA  & 33 &  32.12  &  72.56  &  79.80  &  41.32  &  37.32  \\
MNLI MM  &  BSA  & 33 &  31.32  &  73.64  &  82.16  &  42.36  &  36.96  \\
SNLI  &  BSA & 50  &  33.88  &  79.28  &  87.08  &  37.00  &  35.40  \\
StoryCloze  &  BSA  & 50 &  50.51  &  84.13  &  81.67  &  82.31  &  64.56  \\
CoPA  &  BSA  & 50 &  34.20  &  49.00  &  49.40  &  38.60  &  30.20  \\
SQuAD  &  F1  & 50 &  8.29  &  74.75  &  62.79  &  79.25  &  61.89  \\
BoolQ  &  BSA  & 50 &  41.84  &  82.20  &  76.12  &  65.56  &  63.72  \\
Winogrande  &  BSA  & 50 &  48.38  &  62.51  &  66.54  &  53.35  &  51.22  \\
WSC273 (OD)  &  BSA  & 50 &  50.55  &  61.90  &  61.54  &  57.51  &  56.78  \\
\hline
NLG  &    &    &    &    &    &    \\
\hline
DART  &  Rouge1  & -- &  3.32  &  67.51  &  19.55  &  62.38  &  16.93  \\
CommonGen  &  Rouge1  & --&  7.15  &  43.07  &  43.85  &  42.05  &  40.36  \\
WMT-16 EN/DE  &  BLEU  & --&  2.74  &  32.94  &  33.05  &  12.46  &  24.91  \\
WMT-16 EN/RO   &  BLEU  & --&  5.26  &  32.00  &  30.24  &  18.78  &  23.59  \\
Gigaword (OD)  &  Rouge1  & --&  0.36  &  21.13  &  14.78  &  31.29  &  15.29  \\
\hline
BB - Emergent  &    &    &    &    &    &    \\
\hline
SocialIQA  &  BSA  & 33  &  38.70  &  61.05  &  57.36  &  64.66  &  49.48  \\
LogicalDeduction  &  BSA  & 20  &  19.08  &  23.58  &  21.50  &  30.00  &  21.58  \\
Strange Stories  &  BSA  & 50  &  37.86  &  51.43  &  41.43  &  54.29  &  36.43  \\
\hline
BB - Not Emergent  &    &    &    &    &    &    \\
\hline
Tracking (7)  &  BSA  & 14  &  14.50  &  13.57  &  14.07  &  13.43  &  15.00  \\
GSM8K  &  EMA  &  25  & 0.68  &  51.87  &  48.76  &  58.46  &  53.31  \\
Causal Judgement  &  BSA  & 50  &  46.05  &  47.37  &  48.03  &  50.00  &  46.71  \\
\hline
\end{tabular}
\caption{Full main experiment results for Llama-2 13B models, seed 10203.}
\label{app:hyp1_llama13B_seed3}
\end{table*}

\begin{table*}[htb]
\footnotesize
\centering
\begin{tabular}{rccccccc}
\hline
\textbf{Test Task}                        & \textbf{Metric} & \textbf{Random} & \textbf{Base (D)} & \textbf{FLAN (D)} & \textbf{SampleGen (D)} & \textbf{Gold (P)} & \textbf{SampleGen (P)} \\
\hline
NLU &  &  &  &  &  &  \\
\hline
QNLI  &  BSA  & 50  &  50.56  &  67.48  &  61.12  &  52.32  &  53.32  \\
MNLI M  &  BSA  & 33  &  31.48  &  63.64  &  72.44  &  41.84  &  39.48  \\
MNLI MM  &  BSA  & 33  &  30.92  &  66.88  &  75.48  &  41.20  &  39.48  \\
SNLI  &  BSA  & 50  &  33.88  &  67.28  &  81.32  &  44.48  &  38.44  \\
StoryCloze  &  BSA  & 50  &  51.74  &  75.57  &  62.32  &  73.60  &  55.00  \\
CoPA  &  BSA  & 50  &  42.40  &  51.80 &  49.40  &  35.20  &  36.60  \\
SQuAD  &  F1  & 50  &  17.39  &  50.44  &  45.41  &  73.94  &  44.39  \\
BoolQ  &  BSA  & 50  &  39.80  &  78.92  &  70.76  &  59.40  &  61.96  \\
Winogrande  &  BSA  & 50   &  50.99  &  52.88  &  57.22  &  50.75  &  48.78  \\
WSC273 (OD)  &  BSA  & 50  &  53.11  &  56.41  &  51.28  &  56.41  &  54.21  \\
\hline
NLG &    &    &    &    &    &    \\
\hline
DART  &  Rouge1  & --  &  4.97  &  64.96  &  52.53  &  59.86  &  41.48  \\
CG  &  Rouge1  & -- &  5.99  &  42.92  &  43.48  &  40.21  &  39.84  \\
WMT-16 EN/DE  &  BLEU  & -- &  4.77  &  30.06  &  29.00  &  16.16  &  14.28  \\
WMT-16 EN/RO   &  BLEU  & -- &  7.58  &  29.21  &  25.61  &  17.42  &  17.09  \\
Gigaword (OD)  &  Rouge1 & -- &  1.70  &  21.13  &  4.30  &  28.61  &  3.64  \\
\hline
BB - Emergent  &    &    &    &    &    &    \\
\hline
SocialIQA  &  BSA & 33  &  35.72  &  48.00  &  53.49  &  51.61  &  48.97  \\
LogicalDeduction  &  BSA  & 20 &  19.00  &  20.83  &  20.92  &  22.33  &  18.42  \\
Strange Stories  &  BSA  & 50 &  45.00  &  42.14  &  44.29  &  45.00  &  34.29  \\
\hline
BB - Not Emergent  &    &    &    &    &    &    \\
\hline
Tracking (7)  &  BSA  & 14 &  14.71  &  14.14  &  14.86  &  11.93  &  14.57  \\
GSM8K  &  EMA  & 25 &  0.00  &  39.28  &  19.93  &  49.61  &  23.48  \\
Causal Judgement  &  BSA  & 50 &  46.05  &  46.71  &  45.39  &  50.66  &  46.05  \\
\hline
\end{tabular}
\caption{Full main experiment results for Llama-2 7B models, seed 42.}
\label{app:hyp1_llama7B}
\end{table*}

\begin{table*}[htb]
\footnotesize
\centering
\begin{tabular}{rccccccc}
\hline
\textbf{Test Task}                        & \textbf{Metric} & \textbf{Random} & \textbf{Base (D)} & \textbf{FLAN (D)} & \textbf{SampleGen (D)} & \textbf{Gold (P)} & \textbf{SampleGen (P)} \\
\hline
NLU &  &  &  &  &  &  \\
\hline
QNLI  &  BSA  & 50  &  50.56  &  51.12  &  63.12  &  54.70  &  51.08  \\
MNLI M  &  BSA  & 33  &  31.96  &  37.20  &  82.56  &  38.88  &  37.24  \\
MNLI MM  &  BSA & 33   &  31.36  &  38.76  &  85.96  &  39.76  &  36.56  \\
SNLI  &  BSA  & 50  &  33.68  &  25.36  &  88.76  &  42.82  &  36.20  \\
StoryCloze  &  BSA  & 50  &  55.32  &  63.44  &  76.86  &  86.10  &  66.70  \\
CoPA  &  BSA  & 50  &  44.40  &  56.00  &  45.80  &  48.80  &  37.40  \\
SQuAD  &  F1  & 50  &  30.83  &  33.31  &  65.65  &  78.14  &  59.44  \\
BoolQ  &  BSA  & 50  &  43.00  &  38.12  &  80.32  &  66.15  &  63.48  \\
Winogrande  &  BSA  & 50  &  51.14  &  53.91  &  71.67  &  55.64  &  51.78  \\
WSC273 (OD)  &  BSA  & 50  &  50.18  &  54.95  &  63.74  &  56.04  &  58.24  \\
\hline
NLG &  &  &  &  &  &  \\
\hline
DART  &  Rouge1  & --  &  20.86  &  25.58  &  61.08  &  63.72  &  53.32  \\
CG  &  Rouge1  & -- &  14.03  &  28.75  &  44.61  &  41.25  &  42.30  \\
WMT-16 EN/DE  &  BLEU  & -- &  4.59  &  5.61  &  32.30  &  11.72  &  23.48  \\
WMT-16 EN/RO   &  BLEU  & -- &  5.87  &  4.40  &  29.04  &  18.44  &  20.06  \\
Gigaword (OD)  &  Rouge1  & -- &  6.08  &  10.48  &  18.01  &  30.95  &  18.12  \\
\hline
BB - Emergent &  &  &  &  &  &  \\
\hline
SocialIQA  &  BSA  & 33  &  33.40  &  63.24  &  64.60  &  64.15  &  50.00  \\
LogicalDeduction  &  BSA  & 20  &  20.92  &  22.58  &  23.33  &  24.42  &  21.83  \\
Strange Stories  &  BSA  & 50  &  33.57  &  57.14  &  45.00  &  52.14  &  48.57  \\
\hline
BB - Not Emergent &  &  &  &  &  &  \\
\hline
Tracking (7)  &  BSA  & 14 &  14.36  &  14.43  &  13.43  &  13.29  &  14.21  \\
GSM8K  &  EMA  & 25 &  29.71  &  1.89  &  53.30  &  62.32  &  58.16  \\
Causal Judgement  &  BSA  & 50 &  46.05  &  46.05  &  46.05  &  55.26  &  45.39  \\
\hline
\end{tabular}
\caption{Full main experiment results for Mistral 7B models, seed 42.}
\label{app:hyp1_mistral7B}
\end{table*}

\begin{table*}[!htb]
\footnotesize
\centering
\begin{tabular}{m{0.1\linewidth}p{0.1\linewidth}p{0.1\linewidth}p{0.1\linewidth}p{0.1\linewidth}p{0.1\linewidth}p{0.1\linewidth}p{0.1\linewidth}}
\hline
\textbf{Train On} & \textbf{Test Task} & \textbf{Random} & \textbf{FLAN-D} & \textbf{SampleGen-D} & \textbf{Gold-P Target} & \textbf{Gold-P Train} & \textbf{SampleGen-P} \\
\hline
\multirow{3}*{NLG} & QNLI & 50 & 54.28 & 50.12 & 51.51 & 49.63 & 50.57 \\
& MNLI - M & 33 & 37.59 & 32.17 & 41.53 & 41.93 & 31.75 \\
& MNLI - MM & 33 & 37.05 & 31.45 & 42.89 & 41.95 & 30.61 \\
\hdashline
\multirow{3}*{\makecell[{{p{\linewidth}}}]{Adjacent Cats.}} & QNLI & 50 & 66.33 & 54.16 & 51.51 & 49.73 & 49.67 \\
& MNLI - M & 33 & 51.20 & 52.77 & 41.53 & 42.75 & 38.36 \\
& MNLI - MM & 33 & 53.13 & 52.72 & 42.89 & 42.81 & 39.83 \\
\hdashline
\multirow{3}*{\makecell[{{p{\linewidth}}}]{Adjacent Cats. + \\ QNLI}} & QNLI & 50 & 74.43 & 69.17 & 51.51 & 49.87 & 52.59 \\
& MNLI - M & 33 & 56.48 & 55.11 & 41.53 & 42.79 & 40.51 \\
& MNLI - MM & 33 & 58.41 & 56.22 & 42.89 & 42.60 & 40.03 \\
\hdashline
\multirow{3}*{\makecell[{{p{\linewidth}}}]{Adjacent Cats. + \\ MNLI}} &QNLI & 50 & 68.00 & 52.80 & 51.51 & 49.71 & 48.75 \\
& MNLI - M & 33 & 64.80 & 67.47 & 41.53 & 42.37 & 39.27 \\
& MNLI - MM & 33 & 64.99 & 69.33 & 42.89 & 42.79 & 38.47 \\
\hline
\end{tabular}
\caption{Results of task similarity experiments for Llama-2 13B, averaged over 3 seeds. All scores are BERTScore Accuracy.}
\label{tab:hyp2_llama13b_numbers_avg}
\end{table*}

\begin{table*}[!htb]
\footnotesize
\centering
\begin{tabular}{m{0.1\linewidth}p{0.1\linewidth}p{0.1\linewidth}p{0.1\linewidth}p{0.1\linewidth}p{0.1\linewidth}p{0.1\linewidth}p{0.1\linewidth}}
\hline
\textbf{Train On} & \textbf{Test Task} & \textbf{Random} & \textbf{FLAN-D} & \textbf{SampleGen-D} & \textbf{Gold-P Target} & \textbf{Gold-P Train} & \textbf{SampleGen-P} \\
\hline
\multirow{3}*{NLG} & QNLI & 50 & 56.08 & 49.72 & 51.44 & 49.44 & 50.60 \\
& MNLI - M & 33 & 35.00 & 33.44 & 41.68 & 42.52 & 31.96 \\
& MNLI - MM & 33 & 34.00 & 32.52 & 43.12 & 43.64 & 30.84 \\
\hdashline
\multirow{3}*{\makecell[{{p{\linewidth}}}]{Adjacent Cats.}} & QNLI & 50 & 69.40 & 55.84 & 51.44 & 49.80 & 50.32 \\
& MNLI - M & 33 & 48.12 & 52.64 & 41.68 & 42.64 & 37.48 \\
& MNLI - MM & 33 & 49.76 & 50.92 & 43.12 & 43.08 & 39.92 \\
\hdashline
\multirow{3}*{\makecell[{{p{\linewidth}}}]{Adjacent Cats. + \\ QNLI}} & QNLI & 50 & 73.72 & 75.28 & 51.44 & 50.04 & 54.52 \\
& MNLI - M & 33 & 60.08 & 57.96 & 41.68 & 43.20 & 42.52 \\
& MNLI - MM & 33 & 60.88 & 59.31 & 43.12 & 42.16 & 41.48 \\
\hdashline
\multirow{3}*{\makecell[{{p{\linewidth}}}]{Adjacent Cats. + \\ MNLI}} & QNLI & 50 & 69.52 & 51.32 & 51.44 & 49.64 & 49.20 \\
& MNLI - M & 33 & 63.96 & 71.60 & 41.68 & 42.08 & 38.32 \\
& MNLI - MM & 33 & 64.52 & 73.60 & 43.12 & 42.52 & 37.56 \\
\hline
\end{tabular}
\caption{Results of task similarity experiments for Llama-2 13B, seed 42. All scores are BERTScore Accuracy.}
\label{tab:hyp2_llama13b_numbers_seed1}
\end{table*}

\begin{table*}[!htb]
\footnotesize
\centering
\begin{tabular}{m{0.1\linewidth}p{0.1\linewidth}p{0.1\linewidth}p{0.1\linewidth}p{0.1\linewidth}p{0.1\linewidth}p{0.1\linewidth}p{0.1\linewidth}}
\hline
\textbf{Train On} & \textbf{Test Task} & \textbf{Random} & \textbf{FLAN-D} & \textbf{SampleGen-D} & \textbf{Gold-P Target} & \textbf{Gold-P Train} & \textbf{SampleGen-P} \\
\hline
\multirow{3}*{NLG} & QNLI & 50 & 57.28 & 50.32 & 51.68 & 49.72 & 50.60  \\
& MNLI - M & 33 & 35.12 & 31.04 & 41.60 & 41.44 & 31.40  \\
& MNLI - MM & 33 & 34.12 & 30.00 & 43.20 & 41.04 & 29.84  \\
\hdashline
\multirow{3}*{\makecell[{{p{\linewidth}}}]{Adjacent Cats.}} & QNLI & 50 & 69.20 & 56.24 & 51.68 & 49.96 & 49.60  \\
& MNLI - M & 33 & 53.24 & 54.48 & 41.60 & 42.64 & 39.52  \\
& MNLI - MM & 33 & 54.76 & 54.88 & 43.20 & 42.40 & 39.60  \\
\hdashline
\multirow{3}*{\makecell[{{p{\linewidth}}}]{Adjacent Cats. + \\ QNLI}} & QNLI & 50 & 74.56 & 56.04 & 51.68 & 49.92 & 49.72  \\
& MNLI - M & 33 & 55.28 & 54.64 & 41.60 & 42.72 & 39.88  \\
& MNLI - MM & 33 & 57.76 & 55.52 & 43.20 & 42.36 & 39.60  \\
\hdashline
\multirow{3}*{\makecell[{{p{\linewidth}}}]{Adjacent Cats. + \\ MNLI}} & QNLI & 50 & 66.52 & 49.16 & 51.68 & 49.64 & 47.52  \\
& MNLI - M & 33 & 65.28 & 57.68 & 41.60 & 42.08 & 41.60  \\
& MNLI - MM & 33 & 65.84 & 59.68 & 43.20 & 43.04 & 41.08  \\
\hline
\end{tabular}
\caption{Results of task similarity experiments for Llama-2 13B, seed 1234. All scores are BERTScore Accuracy.}
\label{tab:hyp2_llama13b_numbers_seed2}
\end{table*}

\begin{table*}[!htb]
\footnotesize
\centering
\begin{tabular}{m{0.1\linewidth}p{0.1\linewidth}p{0.1\linewidth}p{0.1\linewidth}p{0.1\linewidth}p{0.1\linewidth}p{0.1\linewidth}p{0.1\linewidth}}
\hline
\textbf{Train On} & \textbf{Test Task} & \textbf{Random} & \textbf{FLAN-D} & \textbf{SampleGen-D} & \textbf{Gold-P Target} & \textbf{Gold-P Train} & \textbf{SampleGen-P} \\
\hline
\multirow{3}*{NLG} &  QNLI & 50 & 49.48 & 50.32 & 51.40 & 49.72 & 50.52  \\
&  MNLI - M & 33 & 42.64 & 32.04 & 41.32 & 41.84 & 31.88  \\
&  MNLI - MM & 33 & 43.04 & 31.84 & 42.36 & 41.16 & 31.16  \\
\hdashline
\multirow{3}*{\makecell[{{p{\linewidth}}}]{Adjacent Cats.}} &  QNLI & 50 & 60.40 & 50.40 & 51.40 & 49.44 & 49.08  \\
&  MNLI - M & 33 & 52.24 & 51.20 & 41.32 & 42.96 & 38.08  \\
&  MNLI - MM & 33 & 54.88 & 52.36 & 42.36 & 42.96 & 39.96  \\
\hdashline
\multirow{3}*{\makecell[{{p{\linewidth}}}]{Adjacent Cats. + \\ QNLI}} &  QNLI & 50 & 75.00 & 76.20 & 51.40 & 49.64 & 53.52  \\
&  MNLI - M & 33 & 54.08 & 52.72 & 41.32 & 42.44 & 39.12  \\
&  MNLI - MM & 33 & 56.60 & 53.84 & 42.36 & 43.28 & 39.00  \\
\hdashline
\multirow{3}*{\makecell[{{p{\linewidth}}}]{Adjacent Cats. + \\ MNLI}} &  QNLI & 50 & 67.96 & 57.92 & 51.40 & 49.84 & 49.52  \\
&  MNLI - M & 33 & 65.16 & 73.12 & 41.32 & 42.96 & 37.88  \\
&  MNLI - MM & 33 & 64.60 & 74.72 & 42.36 & 42.80 & 36.76  \\
\hline
\end{tabular}
\caption{Results of task similarity experiments for Llama-2 13B, seed 10203. All scores are BERTScore Accuracy.}
\label{tab:hyp2_llama13b_numbers_seed3}
\end{table*}

\begin{table*}[!htb]
\footnotesize
\centering
\begin{tabular}{m{0.1\linewidth}p{0.1\linewidth}p{0.1\linewidth}p{0.1\linewidth}p{0.1\linewidth}p{0.1\linewidth}p{0.1\linewidth}p{0.1\linewidth}}
\hline
\textbf{Train On} & \textbf{Test Task} & \textbf{Random} & \textbf{FLAN-D} & \textbf{SampleGen-D} & \textbf{Gold-P Target} & \textbf{Gold-P Train} & \textbf{SampleGen-P} \\
\hline
\multirow{3}*{NLG} & QNLI & 50 & 50.84 & 50.12 & 52.32 & 55.08 & 50.52 \\
& MNLI - M & 33 & 37.92 & 32.24 & 41.84 & 34.64 & 31.96 \\
& MNLI - MM & 33 & 36.16 & 32.28 & 41.20 & 33.84 & 33.40 \\
\hdashline
\multirow{3}*{\makecell[{{p{\linewidth}}}]{Adjacent Cats.}} & QNLI & 50 & 59.40 & 54.60 & 52.32 & 52.52 & 53.08 \\
& MNLI - M & 33 & 44.68 & 49.24 & 41.84 & 38.36 & 37.04 \\
& MNLI - MM & 33 & 46.80 & 50.00 & 41.20 & 38.80 & 35.84 \\
\hdashline
\multirow{3}*{\makecell[{{p{\linewidth}}}]{Adjacent Cats. + \\ QNLI}} & QNLI & 50 & 66.16 & 64.76 & 52.32 & 52.16 & 49.56 \\
& MNLI - M & 33 & 49.72 & 47.60 & 41.84 & 38.28 & 36.00 \\
& MNLI - MM & 33 & 50.96 & 49.56 & 41.20 & 39.12 & 35.80 \\
\hdashline
\multirow{3}*{\makecell[{{p{\linewidth}}}]{Adjacent Cats. + \\ MNLI}} & QNLI & 50 & 58.64 & 50.92 & 52.32 & 52.12 & 50.44 \\
& MNLI - M & 33 & 48.76 & 58.84 & 41.84 & 37.96 & 40.36 \\
& MNLI - MM & 33 & 49.40 & 62.16 & 41.20 & 36.80 & 40.20 \\
\hline
\end{tabular}
\caption{Results of task similarity experiments for Llama-7B, seed 42. All scores are BERTScore Accuracy.}
\label{tab:hyp2_llama7b_numbers}
\end{table*}

\begin{table*}[!htb]
\footnotesize
\centering
\begin{tabular}{m{0.1\linewidth}p{0.1\linewidth}p{0.1\linewidth}p{0.1\linewidth}p{0.1\linewidth}p{0.1\linewidth}p{0.1\linewidth}p{0.1\linewidth}}
\hline
\textbf{Train On} & \textbf{Test Task} & \textbf{Random} & \textbf{FLAN-D} & \textbf{SampleGen-D} & \textbf{Gold-P Target} & \textbf{Gold-P Train} & \textbf{SampleGen-P} \\
\hline
\multirow{3}*{NLG} & QNLI & 50 & 50.56 & 49.16 & 50.80 & 53.44 & 50.40 \\
& MNLI - M & 33 & 31.64 & 33.16 & 37.28 & 37.48 & 33.32 \\
& MNLI - MM & 33 & 31.40 & 31.56 & 38.64 & 40.64 & 31.48 \\
\hdashline
\multirow{3}*{\makecell[{{p{\linewidth}}}]{Adjacent Cats.}} & QNLI & 50 & 60.40 & 58.76 & 50.80 & 56.08 & 48.92 \\
& MNLI - M & 33 & 41.04 & 55.60 & 37.28 & 38.32 & 38.04 \\
& MNLI - MM & 33 & 44.44 & 58.28 & 38.64 & 37.80 & 37.16 \\
\hdashline
\multirow{3}*{\makecell[{{p{\linewidth}}}]{Adjacent Cats. + \\ QNLI}} & QNLI & 50 & 76.60 & 81.56 & 50.80 & 55.92 & 49.72 \\
& MNLI - M & 33 & 46.32 & 56.88 & 37.28 & 39.20 & 38.56 \\
& MNLI - MM & 33 & 47.80 & 58.04 & 38.64 & 39.68 & 36.80 \\
\hdashline
\multirow{3}*{\makecell[{{p{\linewidth}}}]{Adjacent Cats. + \\ MNLI}} & QNLI & 50 & 52.88 & 51.72 & 50.80 & 56.72 & 48.12 \\
& MNLI - M & 33 & 56.04 & 79.60 & 37.28 & 38.08 & 36.44 \\
& MNLI - MM & 33 & 56.96 & 81.16 & 38.64 & 39.56 & 36.16 \\
\hline
\end{tabular}
\caption{Results of task similarity experiments for Mistral 7B, seed 42. All scores are BERTScore Accuracy.}
\label{tab:hyp2_mistral7b_numbers}
\end{table*}

\subsection{Confounding Factor: Prompt Complexity}
\label{app:full_results_prompt_complexity}
Here we present the full experiment details from our prompt complexity experiments, discussed in Section \ref{sec:main:prompt_complexity}. Figure \ref{fig:prompt-effectiveness} presents the radar chart of our prompt complexity experiments, and Table \ref{app:tab:transl-prompts} presents the numerical results for our prompt translation experiments. Table \ref{app:tab:conv-prompts} presents the full prompts from all complexity levels, as well as the translated prompt. These prompts were generated using ChatGPT-4o. Table \ref{app:tab:conv-prompts-prompts} presents the instructions given to ChatGPT-4o for generating each kind of prompt.

\begin{table*}[!htb]
\footnotesize
\centering
\begin{tabular}{rcccccc}
\\
\toprule
\textbf{Prompt Setting} &  \textbf{Base-D}  &  \textbf{FLAN-D } &  \textbf{SampleGen-D}  &  \textbf{Gold-P}  &  \textbf{SampleGen-P} & \textbf{Random}\\
\midrule
Original &  \textbf{36.93}  &  \textbf{61.13}  &  \textbf{62.18}  &  \textbf{63.71}  &  \textbf{54.11} & 33  \\
Transl. Avg Seeds &  33.50  &  33.63  &  34.07  &  34.35  &  33.29 & 33 \\
Transl. Seed 42  &  32.36  &  35.79  &  34.17  &  33.66  &  34.23 & 33 \\
Transl. Seed 1234  &  34.04  &  31.91  &  33.66  &  34.69  &  31.78 & 33 \\
Transl. Seed 10203  &  34.11  &  33.20  & 34.37   &  34.69  & 33.85  & 33 \\
\bottomrule
\end{tabular}
\caption{Results for our prompt translation experiments on SocialIQA, for Llama-2 13B across 3 seeds. In the translated prompt setting, all base and instruction-tuned models perform at or near the random baseline, even if those models were able to solve SocialIQA with the original English prompt.}
\label{app:tab:transl-prompts}
\end{table*}

\begin{table*}[!htb]
\footnotesize
\centering
\begin{tabular}{p{2cm}p{12cm}}
\\
\toprule
\textbf{Prompt} & \textbf{Prompt Format}\\ 
\midrule
\textbf{Regular} & Question: \{\} Options: \{\} \\ \\
Complexity Level  1& Consider the following inquiry: \{\} The potential selections for your consideration are as follows: \{\} \\  \\
Complexity Level 2 &The following inquiry is presented for your contemplation: \{\} The array of possible responses available for selection is enumerated as follows: \{\} \\ \\
Complexity Level 3 &Presented herewith for your intellectual engagement is the following query: \{\} The spectrum of conceivable responses from which you may elect is delineated as follows: \{\}\\ \\
Complexity Level 4 &For your esteemed deliberation, the subsequent interrogative is posited: \{\} The compendium of potential responses, from which one may judiciously choose, is as follows: \{\} \\ \\
Complexity Level 5 &For your erudite examination and discerning judgment, the following enigma is herein proposed: \{\} The exhaustive enumeration of conceivable answers, from which one might sagaciously select, is detailed as follows: \{\}\\ \\
Complexity Level 6 &For your profound intellectual consideration and meticulous scrutiny, the subsequent conundrum is hereby articulated: \{\} The comprehensive catalog of potential resolutions, from which you may astutely elect, is meticulously delineated as follows: \{\}\\ \\
Translation Scottish Gaelic & Thoir beachd air an rannsachadh a leanas: \{\} Tha na roghainnean a dh’fhaodadh a bhith agad airson do bheachdachadh mar a leanas: \{\}\\
\bottomrule
\end{tabular}
\caption{Prompts from our prompt complexity and prompt translation experiments. All complex prompts and prompt translations were generated using ChatGPT-4o-mini on 05. September, 2024. The corresponding instructions given to ChatGPT-4o-mini to generate these prompts are presented in Table \ref{app:tab:conv-prompts-prompts}.}
\label{app:tab:conv-prompts}
\end{table*}

\begin{table*}[!htb]
\footnotesize
\centering
\begin{tabular}{p{2cm}p{12cm}}
\\
\toprule
\textbf{Setting} & \textbf{Prompt Text for ChatGPT-4o}\\ 
\midrule
Complexity Level 1 & Please rephrase the following question format in a more convoluted way. Be sure to add junk and filler words.

Question: \{question\}? Options: \{options\}. 
\\  \\
Subsequent Levels & Please make it even more convoluted.
\\  \\
Translation Scottish Gaelic & Translate the following text to Scottish Gaelic:

\{\} 
\\ \\
\bottomrule
\end{tabular}
\caption{Instructions for ChatGPT-4o-mini used for generating the prompts in Table \ref{app:tab:conv-prompts}.}
\label{app:tab:conv-prompts-prompts}
\end{table*}

\begin{table*}[!htb]
\footnotesize
\centering
\begin{tabular}{cccccc}
\toprule
 Prompt & BASE (D) & FLAN (D) & SampleGen (D) & Gold Pipeline & SG Pipeline \\
 \midrule
Original Prompt &  \textbf{36.93}	&61.13	&\textbf{62.18}&\textbf{63.71}	&\textbf{54.11}\\
Complexity 1 & 34.95 & \textbf{62.21} & 51.01 & 52.33 & 45.65 \\
Complexity 2 & 33.59 & 61.18 & 46.43 & 52.06 & 39.21 \\
Complexity 3 & 34.37 & 58.46 & 48.32 & 48.56 & 41.04 \\
Complexity 4 & 35.27 & 59.61 & 47.14 & 48.10 & 39.23 \\
Complexity 5 & 34.24 & 57.28 & 44.34 & 46.77 & 38.51 \\
Complexity 6 & 34.37 & 52.75 & 40.03 & 48.00 & 36.22 \\
\hline
\end{tabular}
\caption{Numerical results from our prompt engineering experiments. Experiments were done on Social IQA and the metric used was BERTScore Accuracy, as in our main experiments. Except in one case, model performance for all prompts was lower than the regular prompt, indicating poorer instructional understanding.}
\label{app:tab:conv-prompts-results}
\end{table*}

\subsection{Confounding Factor: Semantic Task Labels}
\label{app:unified_vs_non}

In this section, we present the results from our experiments with semantic task labels. Table \ref{app:tab:results-unified} presents the results, which were obtained from Llama-2 13B on one seed only. Overall, the models tended to perform better in the setting with default task labels.

\begin{table*}[t]
\footnotesize
\begin{center}
\begin{tabular}{p{2cm}p{1cm}p{1cm}p{1cm}p{1cm}p{1cm}p{1cm}p{1cm}p{1cm}p{1cm}p{1cm}}
\\
\toprule
Task  &  Metric  &  Random Baseline  &  BASE-D  &  FLAN-D   &  Sample Gen-D (Unif.) &  Sample Gen-D  &  Gold Pipeline (Unif.)  &  Gold Pipeline &  Sample Gen Pipeline (Unif.)  &  Sample Gen Pipeline \\
\midrule
NLI  &   \\
\hline
QNLI  &  BSA  &  50  &  50.56  &  60.06  &  59.72  &  \textbf{66.04}  &  50.12  &  \textbf{50.72}  &  \textbf{51.20}  &  49.16 \\
SNLI  &  BSA  &  33  &  33.88  &  81.04  &  86.08  &  \textbf{87.64}  &  \textbf{42.60}  &  37.44  &  35.28  &  \textbf{36.72} \\
StoryCloze  &  BSA  &  50  &  50.51  &  81.03  &  \textbf{88.62}  &  76.32  &  61.41  &  \textbf{81.29}  &  58.95  &  \textbf{61.79} \\
CoPA  &  BSA  &  50  &  34.20  &  46.40  &  43.20  &  \textbf{45.00}  &  \textbf{40.80}  &  37.40  &  \textbf{32.20}  &  20.60 \\
SQuAD  &  F1  &  50  &  10.77  &  60.82  &  \textbf{54.31}  &  43.12  &  \textbf{81.53}  &  78.12  &  \textbf{45.64}  &  42.44 \\
BoolQ  &  BSA  &  50  &  41.84  &  82.57  &  75.44  &  \textbf{77.36}  &  57.88  &  \textbf{65.16}  &  63.04  &  \textbf{65.84} \\
Winogrande  &  BSA  &  50  &  48.38  &  63.14  &  64.33  &  \textbf{67.88}  &  50.28  &  \textbf{53.75}  &  50.04  &  \textbf{51.38} \\
WSC273  &  BSA  &  50  &  50.55  &  63.00  &  60.07  &  \textbf{62.64}  &  56.41  &  56.41  &  \textbf{65.57}  &  60.44 \\
\hline
NLU  \\
\hline
DART  &  Rouge1  &  50  &  3.32  &  67.37  &  \textbf{51.59}  &  15.71  &  61.39  &  \textbf{62.50}  &  \textbf{39.47}  &  12.40 \\
Common Gen  &  Rouge1  &  50  &  7.15  &  43.42  &  \textbf{43.26}  &  42.42  &  40.96  &  \textbf{41.45}  &  37.84  &  \textbf{38.87} \\
WMT EN/DE  &  BLEU  &  50  &  2.74  &  32.98  &  33.49  &  \textbf{33.63}  &  \textbf{16.35}  &  12.73  &  25.01  &  \textbf{25.29} \\
WMT EN/RO   &  BLEU  &  50  &  5.26  &  32.39  &  \textbf{31.08}  &  27.54  &  12.20  &  \textbf{19.02}  &  \textbf{20.84}  &  18.36 \\
Gigaword (OD)  &  Rouge1  &  50  &  0.36  &  21.43  &  \textbf{16.15}  &  11.94  &  18.43  &  \textbf{31.49}  &  \textbf{13.38}  &  12.32 \\
\hline
BB + GSM8K \\
\hline
SocialIQA  &  BSA  &  33  &  38.70  &  60.92  &  64.34  &  \textbf{65.37}  &  53.94  &  \textbf{62.98}  &  48.32  &  \textbf{57.30} \\
Logical D.  &  BSA  &  20  &  19.08  &  27.75  &  \textbf{26.50}  &  22.25  &  23.58  &  \textbf{28.42}  &  20.33  &  \textbf{21.08} \\
Strange Stories  &  BSA  &  50  &  37.86  &  50.00  &  43.57  &  \textbf{47.86 } &  \textbf{51.43}  &  48.57  &  \textbf{44.29}  &  41.43 \\
Tracking (7)  &  BSA  &  14  &  14.50  &  13.21  &  13.93  &  \textbf{15.21}  &  12.21  &  \textbf{13.86}  &  14.71  &  \textbf{15.36} \\
GSM8K  &  EMA  &  25  &  0.68  &  51.03  &  \textbf{54.97}  &  49.44  &  45.82  &  \textbf{57.85}  &  45.85  &  \textbf{53.37} \\
Causal J.  &  BSA  &  50  &  46.05  &  46.05  &  46.05  &  \textbf{49.34}  &  \textbf{53.95}  &  49.34  &  46.05  &  \textbf{47.37} \\
\hline
\end{tabular}
\caption{Numerical results from our semantic label experiments, conducted on Llama 2 13B, on one seed only (42). `Unif.' indicates the evaluation setups with ``unified'' (i.e., semantically-neutral) prompt formats. As in the main results, we contrast between direct evaluation setups (D), where a model is tested directly on the task, and pipeline setups (P), where a base model is tested while being given in-context samples either from the training set (Gold Pipeline) or samples generated by SampleGen (SampleGen Pipeline). Overall, the non-unified models outperform the unified models.}
\label{app:tab:results-unified}
\end{center}
\end{table*}

\subsection{Confounding Factor: Instruction-Tuning Tasks}
\label{app:task_ablation_experiments}

In this section, we present the results demonstrating generalisability for our the experiments on instruction-tuning tasks, discussed in Section \ref{subsec:instruction-tuning-tasks}. In addition to Llama-2 13B, the aggregated results for which were presented in the main body of the paper, we experiment with Llama-2 7B and Mistral 7B. The experiments for the latter two models were done on one seed only (42).

Full numerical results for Llama-2 13B, averaged across 3 seeds and for each individual seed, are presented in Tables \ref{tab:hyp2_llama13b_numbers_avg} - \ref{tab:hyp2_llama13b_numbers_seed3}. We present the results for Llama-2 7B and Mistral 7B on one seed in Tables \ref{tab:hyp2_llama7b_numbers} and \ref{tab:hyp2_mistral7b_numbers}, respectively.

\begin{figure}[!htb]
    \centering
    \includegraphics[width=\linewidth,trim={1.5cm 0cm 0cm 5.5cm},clip]{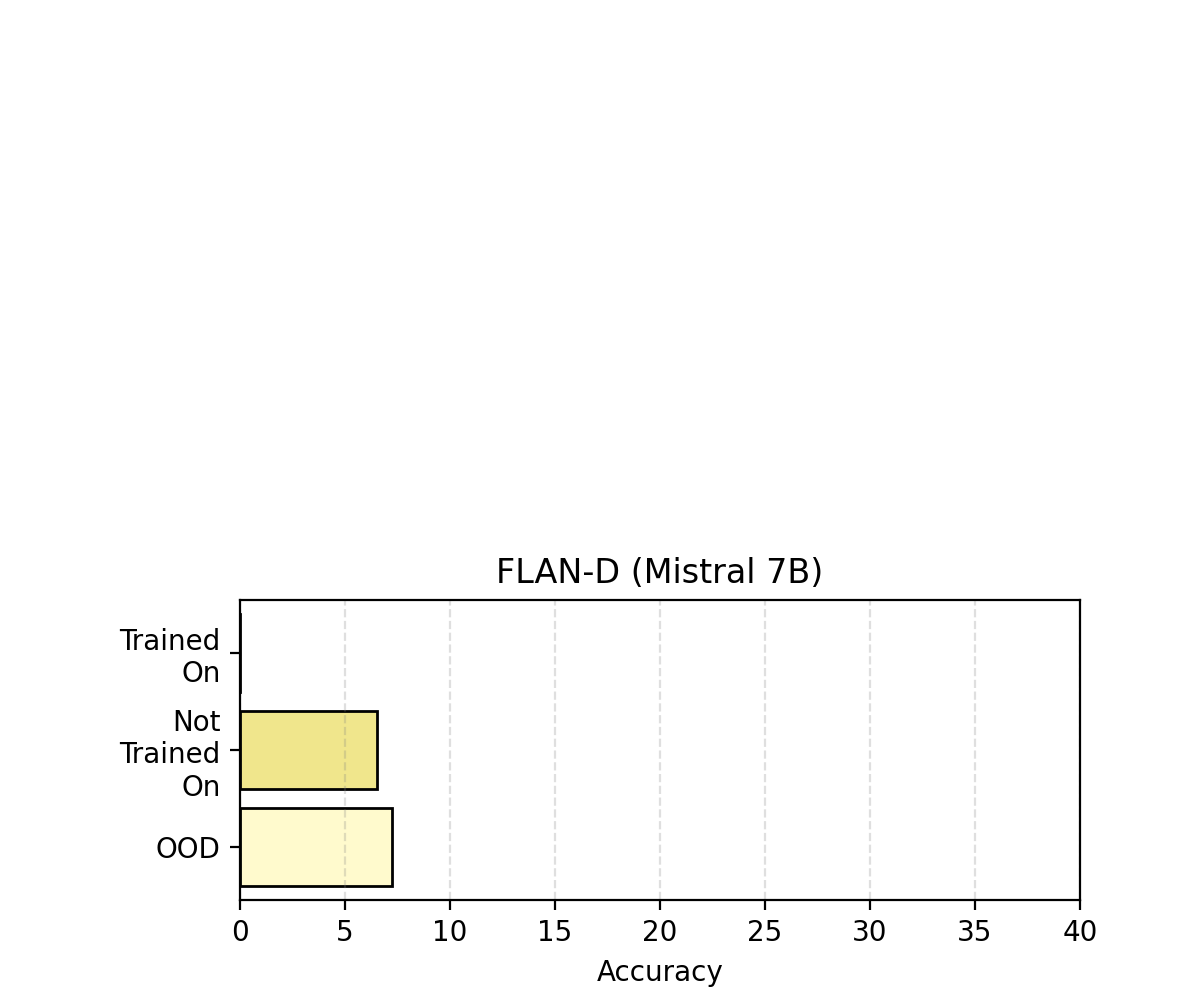}
    \caption{Overall performance trend of our Mistral 7B FLAN model across three clusters of tasks, trained and evaluated on one seed (42). \textbf{These results are not included in our analysis due to the unexpected output formats which were difficult to evaluate, as discussed in Appendix \ref{app:task_ablation_experiments}.}
    }
    \label{fig:app:normalized_bars_mistral7b}
\end{figure}

Figure \ref{fig:app:normalized_bars_llama7b} shows the aggregated scores for Llama-2 7B, which demonstrating a similar trend as for Llama-2 13B and confirm our hypothesis. However, these results do not generalise to Mistral 7B, as this model was more prone to generating extraneous text in its responses, as shown in Table \ref{app:tab:mistral-error-analysis}. This made it difficult to determine the best answer choice match for the model's response during BERTScore Accuracy calculation. Additionally, Mistral displayed a tendency to designate its choices by means of emojis (e.g. printing all answer options with a check mark emoji next to the option it deemed correct). This resulted in all answer options being printed, which similarly complicated the evaluation. These factors resulted in poorer overall performance for Mistral, as shown in Figure \ref{fig:app:normalized_bars_mistral7b}.

\begin{table*}[!htb]
    \centering
    \begin{tabular}{cc}
    \includegraphics[width=0.5\linewidth,trim={1.5cm 0cm 0cm 5.5cm},clip]{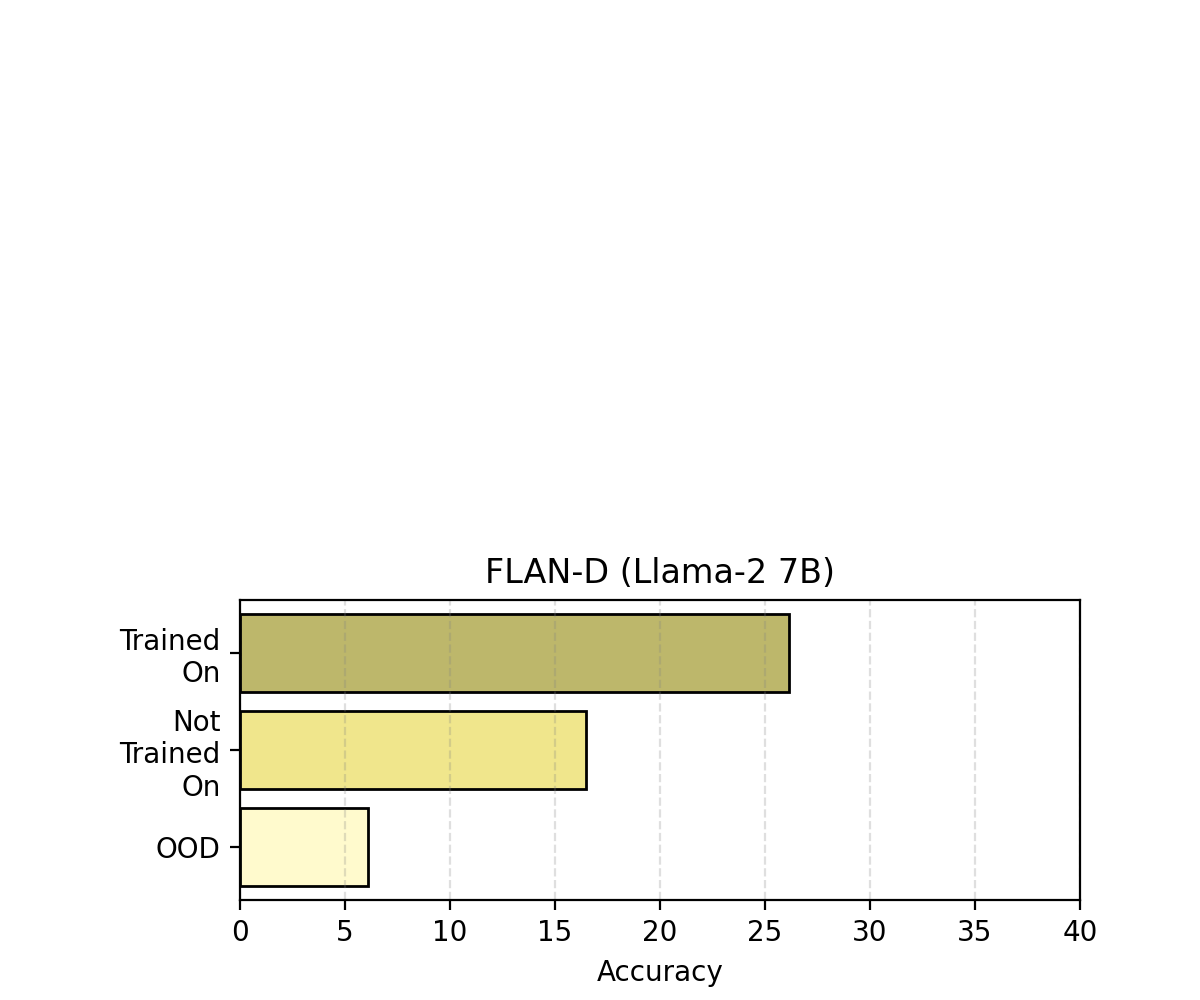} &
    \includegraphics[width=0.5\linewidth,trim={1.5cm 0cm 0cm 5.5cm},clip]{figure_FLAN-D.png} \\
    \end{tabular}
        \caption{Overall performance trend of our Llama-2 7B FLAN model (left) across three clusters of tasks, trained and evaluated on one seed (42). The Llama-2 13B model, with standard deviations across 3 seeds, is shown on the right for comparison. As with the 13B model, we observe that the scores of the 7B model on the in-distribution tasks are higher than those on the out-of-distribution tasks, strengthening our finding that instruction-tuned models are more adept at the former.}
    \label{fig:app:normalized_bars_llama7b}
\end{table*}

\begin{table*}[!htb]
\footnotesize
\centering
\begin{tabular}{p{2cm}p{5cm}p{5cm}p{1cm}}
\\
\toprule
\textbf{Task} &\textbf{Input} & \textbf{Raw Output (FLAN-D)} & \textbf{Target}\\ 
\midrule
SNLI & \makecell[{{p{\linewidth}}}]{Does the sentence "Luther sympathised with some of the peasants' grievances, as he showed in his response to the Twelve Articles in May 1525, but he reminded the aggrieved to obey the temporal authorities." answer the question "How did Luther express the destruction?"\\  \\ yes; no} & \makecell[{{p{\linewidth}}}]{Answer Solution: no\&nbsp; \&nbsp; \&nbsp; \&nbsp; no\&nbsp; \&nbsp; \&nbsp; \&nbsp; \&nbsp; \&nbsp; \&nbsp; \&nbsp; \&nbsp; \&nbsp;\&nbsp; \&nbsp; \&nbsp;
\\ \\ \textit{(... pattern continues)}} & no \\ \\ \hdashline \\
GSM8K & \makecell[{{p{\linewidth}}}]{Question: Mia and Emma are currently 16 years apart in age. If Mia, who is younger than Emma, is 40 years old, what's the average of their ages?\\  \\ "40", "48", "96", "56"} & \makecell[{{p{\linewidth}}}]{Answer Solution: "48" 40 + 48 = 88 88 ÷ 2 = 44 44 + 40 = 84 84 ÷ 2 = 42 42 + 40 = 82 82 ÷ 2 = 41 41 + 40 = 81 81 ÷ 2 = 40.5 40.5...  \\ \\ \textit{(... pattern continues)}} & 48 \\ \\ \hdashline \\
Winogrande & \makecell[{{p{\linewidth}}}]{
Write the next sentence. \\ \\ The living room was cleaner than the kitchen, because we spend more time in the the \_ .\\  \\ living room; kitchen} & \makecell[{{p{\linewidth}}}]{Answer Solution: living room \textit{(check emoji)}; kitchen \textit{(x-mark emoji)}} & kitchen \\ \\
\bottomrule
\end{tabular}
\caption{Example outputs from FLAN-D Mistral-7B, for the experiments discussed in Appendix \ref{app:task_ablation_experiments}. We note that Mistral FLAN has a tendency to generate extraneous text not relevant to its answer choice, making it difficult to determine the best answer choice match during BERTScore calculation. Additionally, Mistral has a tendency to express choices as emojis (e.g. check mark vs. x-mark), which results in both answer options being printed and similarly complicates the evaluation.}
\label{app:tab:mistral-error-analysis}
\end{table*}

\section{Instruction-Tuning Prompts}
\label{app:prompts}

In this section, we provide the full instruction-tuning prompts used for training FLAN and SampleGen. System prompts and wrapper tokens such as \textit{\textless/s\textgreater} are included.
\begin{table*}[!htb]
\footnotesize
\begin{tabular}{p{0.2\linewidth}p{0.7\linewidth} }
\toprule
\textbf{Model} & \textbf{Instruction-Tuning Prompt}\\ \hline \\
FLAN  & \makecell[{{p{\linewidth}}}]{\textless s\textgreater[INST] \textless \textless SYS\textgreater \textgreater \\ You are a large language model for solving tasks. To solve a task, you print the answer using the following format: 'Answer Solution: ...'\textless \textless /SYS\textgreater \textgreater \\  \\ Continue the following story. \\  \\ The baseball game cost more for us to go to than the football game because the \_ was professional. \\  \\ football game; baseball game[/INST] \\ Answer Solution: baseball game\textless/s\textgreater
}      \\ \\
\hline \\
SampleGen  & \makecell[{{p{\linewidth}}}]{\textless s\textgreater[INST] \textless\textless SYS\textgreater\textgreater You are a large language model for generating examples. To answer a question, you return a list of samples and use them to answer the question provided. The format of the output is \enquote{Answer Solution: ... END\_EXAMPLES} \textless\textless/SYS\textgreater\textgreater \\ \\ Continue writing the following text.  \\ Jennifer wanted Lindsey to drink the water even though \_ wasn't feeling well and had an upset stomach.  \\ Jennifer; Lindsey[/INST]  \\ \\ Example 1:  \\ Context 1: Since Cynthia was a morning person while Lindsey was a night person, \_ tended to wake up very late on weekends.  \\ Options 1: Cynthia; Lindsey  \\ Answer 1: Lindsey  \\  \\ Example 2:  \\ Context 2: Nelson was able to shoot the deer easily even though Ian was closer because \_ had good aim.  \\ Options 2: Nelson; Ian  \\ Answer 2: Nelson   \\ \\ Solution:  \\ Context Solution: Jennifer wanted Lindsey to drink the water even though \_ wasn't feeling well and had an upset stomach.  \\ Options Solution: Jennifer; Lindsey  \\ Answer Solution: Lindsey  \\ END\_EXAMPLES \\ \textless/s\textgreater } 
\\ \\
\hline
\end{tabular}
\caption{\label{tab:training_prompts} Instruction-tuning prompt formats for SampleGen and FLAN. The system prompt is included.}
\end{table*}

\begin{table*}[htp]
\footnotesize
\renewcommand{\arraystretch}{1.5}
\begin{center}
\begin{tabular}{ | m{3cm} | m{5cm}| m{3cm} | }
\hline
Model & Instruction Tuning Tasks & Evaluation Tasks  \\
\hline
NLG & All NLG Tasks & \multirow{4}{2cm}{QNLI and MNLI from the Language Inference Cluster} \\
NLG+(NLU-NLI) & All NLG tasks, and all NLU tasks except for those in the NLI cluster & \\
NLG+(NLU-NLI+QNLI) & All NLG tasks, and all NLU tasks except for those in the NLI cluster and QNLI & \\ 
NLG+(NLU-NLI+MNLI) & All NLG tasks, and all NLU tasks except for those in the NLI cluster and MNLI & \\ 
\hline
\end{tabular}
\end{center}
\caption{\label{tab:hyp2evaltypes} Descriptions of the models used in our supplementary task similarity experiments.
}
\end{table*}

\end{document}